\documentclass{article}
\usepackage[table]{xcolor}
\usepackage{iclr2027_conference,times}

\usepackage{amsmath,amsfonts,bm}

\def\eqref#1{equation~\ref{#1}}

\def\1{\bm{1}}

\DeclareMathAlphabet{\mathsfit}{\encodingdefault}{\sfdefault}{m}{sl}
\SetMathAlphabet{\mathsfit}{bold}{\encodingdefault}{\sfdefault}{bx}{n}

\newcommand{\E}{\mathbb{E}}

\newcommand{\KL}{D_{\mathrm{KL}}}

\definecolor{refred}{HTML}{A64545}
\definecolor{citeblue}{HTML}{4477AA}
\usepackage[
    colorlinks=true,
    linkcolor=refred,
    citecolor=citeblue,
    urlcolor=citeblue
]{hyperref}
\usepackage{url}
\usepackage{booktabs}
\usepackage{multirow}
\usepackage{amsmath}
\usepackage{amssymb}
\usepackage{graphicx}
\usepackage{algorithm}
\usepackage{algpseudocode}
\usepackage{enumitem}
\usepackage{wrapfig}
\usepackage{soul}

\usepackage{fontawesome5}
\definecolor{cbest}{RGB}{250,214,214}
\definecolor{csecond}{RGB}{253,238,238}
\newcommand{\bst}[1]{\cellcolor{cbest}\textbf{#1}}
\newcommand{\snd}[1]{\cellcolor{csecond}\underline{#1}}
\newsavebox{\fitbox}
\newcommand{\fit}[1]{\sbox{\fitbox}{#1}%
  \ifdim\wd\fitbox>\linewidth\resizebox{\linewidth}{!}{\usebox{\fitbox}}\else\usebox{\fitbox}\fi}
\newcommand{\hd}[1]{\begin{tabular}[c]{@{}c@{}}#1\end{tabular}}

\newcommand{\bestmark}{\colorbox{cbest}{\textbf{Best}}}
\newcommand{\secondmark}{\colorbox{csecond}{\underline{second}}}

\definecolor{gain}{rgb}{0.0,0.45,0.15}
\newcommand{\gain}[1]{{\color{gain}#1}}

\newcommand{\placeholder}{\textnormal{--}}

\title{Omni-Diffusion-Distill: Few-Step Distillation of Unified Multimodal Diffusion Large Language Models}

\newcommand{\projecturl}{https://hongh0.github.io/omni-diffusion-distill}
\newcommand{\codeurl}{https://CODE-URL}
\newcommand{\modelurl}{https://MODEL-URL}
\newcommand{\corrmark}{\textsuperscript{\,\faEnvelope[regular]}}

\author{\hspace*{-\tabcolsep}%
\begin{minipage}{\textwidth}
\centering
\normalfont
{\large
\textbf{Hong Huang}$^{1,2*}$\quad
\textbf{Chenhongyi Yang}$^{2}$\quad
\textbf{Junzhe Sun}$^{2}$\quad
\textbf{Animesh Sinha}$^{2}$
\\[0.45em]
\textbf{Wuyang Chen}$^{1}$\corrmark\quad
\textbf{Yifan Jiang}$^{2\dagger}$\corrmark
}
\\[0.8em]
{\small
$^{1}$Simon Fraser University
\quad
$^{2}$Meta
}
\\[0.75em]
\href{\projecturl}{\textcolor{blue!65!black}{\faGlobe\ \textbf{Project Page}}}
\end{minipage}%
\hspace*{-\tabcolsep}}

\iclrfinalcopy
\begin{document}

\maketitle
\begingroup
\renewcommand{\thefootnote}{}
\footnotetext{$^{*}$Work done during an internship at Meta.}
\footnotetext{$^{\dagger}$Work done during full-time employment at Meta.}
\footnotetext{\textsuperscript{\faEnvelope[regular]}\,Corresponding authors.}
\endgroup
\lhead{Preprint}

\begin{figure}[h!]
\vspace{-2em}
\includegraphics[width=\linewidth]{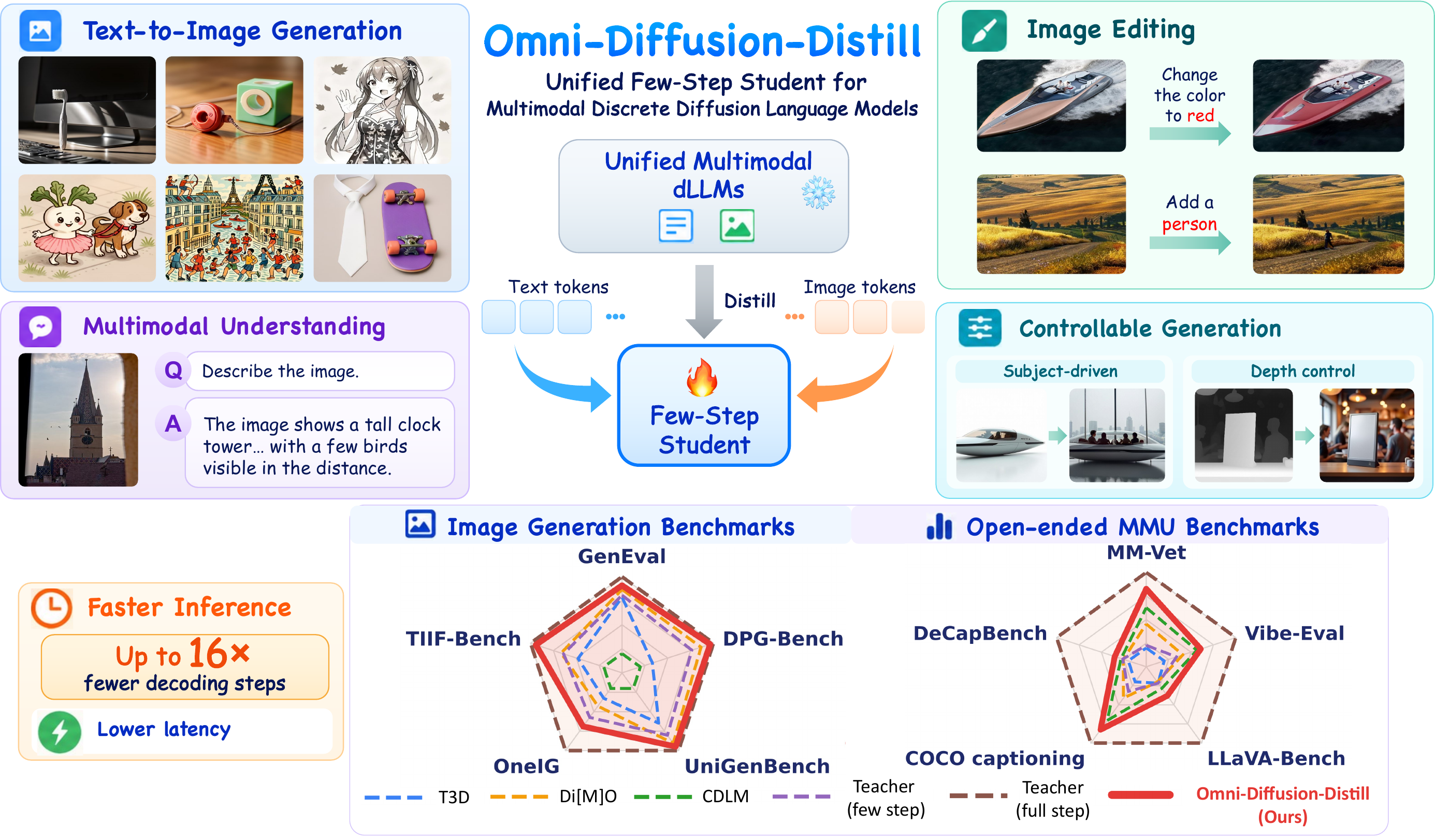}
\vspace{-2em}
\caption{\textbf{Few-step distillation of a unified multimodal dLLM.} With up to $16\times$ fewer
decoding steps, our student dLLM unifies four tasks: text-to-image generation, image editing,
controllable generation and multimodal understanding.}
\label{fig:teaser}
\end{figure}
\begin{abstract}

Unified multimodal diffusion large language models (dLLMs) offer a single architecture for both image generation and multimodal understanding, but their iterative decoding requires tens to hundreds of forward passes.
Existing few-step distillation methods largely focus on either image generation or text generation, making it unclear how to compress a fully discrete multimodal dLLM into a single efficient student while preserving both generation and understanding.
We introduce \textbf{Omni-Diffusion-Distill}, a unified two-stage distillation framework that
retains strong generation and understanding capabilities while substantially reducing the inference cost of a unified multimodal dLLM.
Omni-Diffusion-Distill aligns the distillation of both generation and understanding, for both images and text, in the discrete token space.
In the first stage, the student is trained to skip decoding steps by replaying cached teacher trajectories,
and in the second stage the student is refined on intermediate states along its own rollouts.
We further remedy two sources of degradation in unified distillation with a pairwise collision penalty that reduces repetition under parallel text decoding,
and entropy-matched guidance that prevents entropy collapse caused by fitting the sharpened teacher distribution in image generation.
Omni-Diffusion-Distill achieves state-of-the-art trade-offs between decoding efficiency and generation and understanding performance for multimodal dLLMs, reducing image generation from $128$ to $8$ decoding steps and multimodal understanding from $512$ to $64$, giving $\textbf{18.2}\times$ and $\textbf{21.2}\times$ wall-clock speedups.
Under these budgets, it scores \textbf{0.828} on GenEval and \textbf{83.0} on DPG-Bench for text-to-image generation, while reaching GPT judge scores of \textbf{20.0} on MM-Vet and \textbf{57.2} on COCO captioning (twice the teacher's 28.4 at the same steps) for multimodal understanding.

\end{abstract}

\section{Introduction}
\label{sec:intro}

Building a single model that can both understand and generate across text and images has become an important direction in multimodal learning.
\emph{Diffusion large language models (dLLMs)} provide a natural foundation for this goal: they generate by iteratively decoding masked tokens with a bidirectional model~\citep{austin2021d3pm,sahoo2024mdlm,nie2025llada}, while vector-quantized image tokenizers allow images to be represented in the same discrete token space as text~\citep{vandenoord2017neural,esser2021taming}.
Recent \emph{unified multimodal dLLMs} exploit this formulation to support both image generation and multimodal understanding within a single network~\citep{yang2025mmada,xin2025luminadimoo}. Their flexibility, however, comes with a substantial inference cost: producing one output can require tens to hundreds of sequential decoding steps.
\emph{Distillation} is the standard way to shorten an iterative sampler, and for a unified model it has to shorten the decoding of images and of text at once.

\begin{figure}[!b]
\centering
\vspace{-1.5em}
\includegraphics[width=\linewidth]{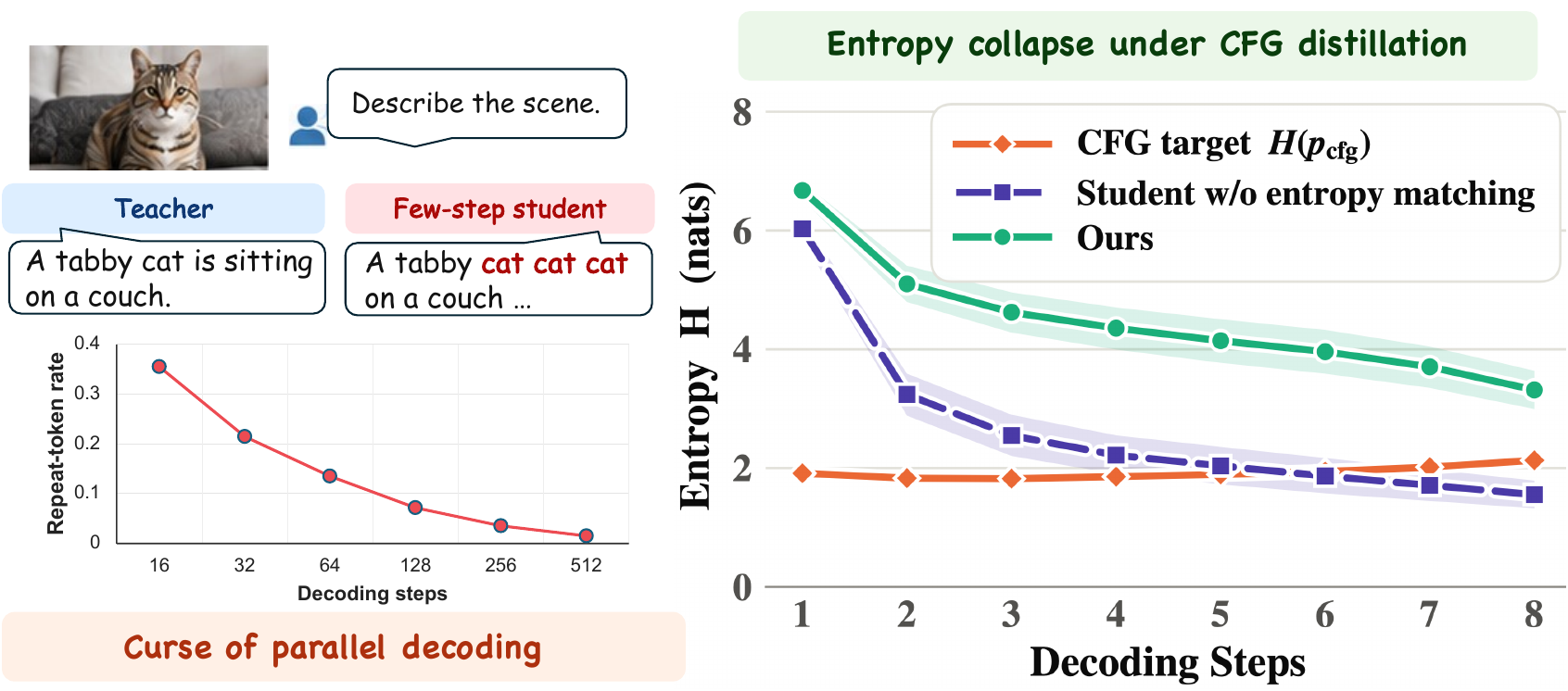}
\vspace{-1.5em}
\caption{\textbf{Challenges in few-step distillation of unified multimodal dLLMs.}
\textbf{Left:} Decoding more text tokens in parallel ignores their dependencies and increases repetition.
\textbf{Right:} A student distilled toward the low-entropy CFG target (\textcolor[HTML]{EB6834}{orange}) keeps losing entropy and ends below it (\textcolor[HTML]{4A3AA7}{purple}). Entropy-matched guidance keeps our student's entropy high (\textcolor[HTML]{1BAF7A}{green}).
}
\vspace{-1.5em}
\label{fig:motivation}
\end{figure}

However, existing few-step distillation methods largely treat image generation~\citep{yin2024dmd,song2023consistency,zhu2025dimo} and text dLLMs~\citep{kim2025cdlm,deschenaux2024beyond,zhang2026t3d} separately, and directly applying them to a unified model raises two modality-specific challenges, illustrated in Figure~\ref{fig:motivation}.
\emph{First}, for text, reducing the number of decoding steps requires committing more tokens in parallel. Tokens committed in the same decoding step are sampled from independent per-position predictions and therefore cannot condition on one another, leading to the \emph{curse of parallel decoding}~\citep{wu2025fastdllm,liu2025copula,kang2026parallelbench}. This problem becomes increasingly severe under aggressive step reduction and often manifests as repetition in long-form generation. 
\emph{Second}, for image generation, removing the unconditional forward pass at inference requires the student to internalize classifier-free guidance (CFG)~\citep{ho2022classifierfree} during distillation. CFG sharpens the teacher distribution throughout decoding~\citep{sanchez2023stay},
and repeatedly matching these sharpened targets progressively lowers the entropy of the student's own predictions. This loss of entropy reduces sample diversity~\citep{ho2022classifierfree,yin2024dmd}. In image editing, it makes the student tend to leave the source image unchanged~\citep{brooks2023instructpix2pix}. Therefore, few-step distillation of a unified dLLM requires not only a common training framework across modalities, but also mechanisms tailored to the distinct challenges introduced by their decoding processes. This raises a core question:

\begin{center}
\vspace{-0.5em}
\fbox{\parbox{0.94\linewidth}{
\centering
\textit{Can we unify the distillation of image and text generation in a multimodal dLLM?}}}
\vspace{-0.5em}
\end{center}

\begin{figure}[t]
\vspace{-2.5em}
\includegraphics[width=\linewidth]{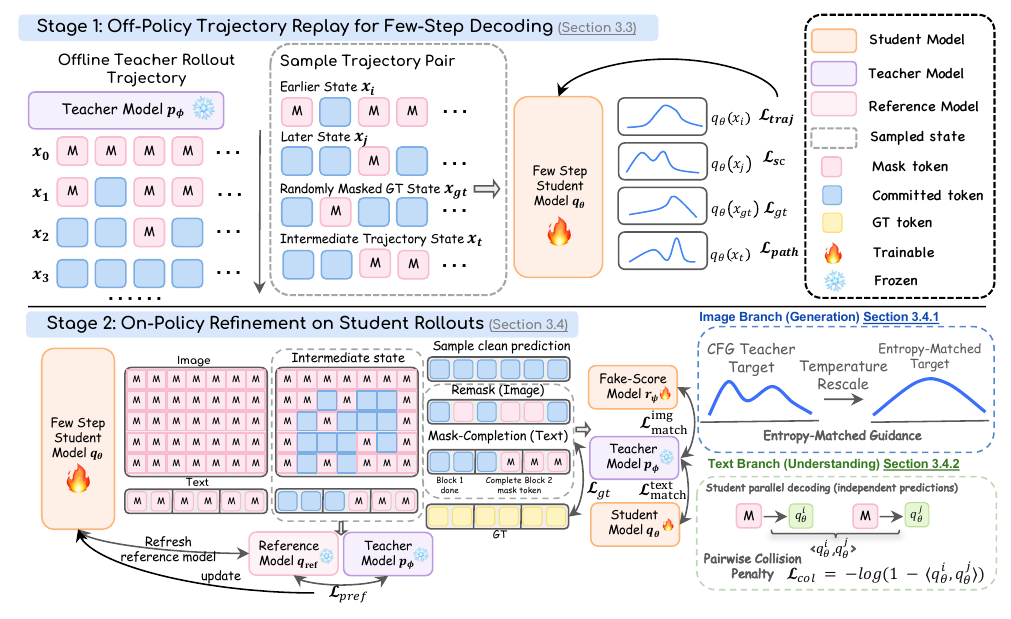}
\vspace{-2.5em}
\caption{\textbf{Overview of Omni-Diffusion-Distill.}
Stage 1 replays cached teacher trajectories off-policy, and Stage 2 refines the student on its own rollouts. Entropy-matched guidance rescales the image teacher's entropy, and a pairwise collision penalty reduces text repetition.
}
\vspace{-1.7em}
\label{fig:pipeline}
\end{figure}

We answer this question with \textbf{Omni-Diffusion-Distill}, a unified few-step distillation framework for
fully discrete multimodal dLLMs. 
In these models, a single denoiser predicts a categorical distribution over tokens for both text and images. 
Image and text decoding can therefore share one distillation framework.
We train Omni-Diffusion-Distill in two stages.
\emph{First}, an off-policy stage replays cached teacher trajectories from which the student learns to
skip decoding steps.
\emph{Second}, an on-policy stage refines the student on intermediate states along its own rollouts.
Importantly, we address the degradation of unified distillation with two remedies:
a \emph{pairwise collision penalty} that penalizes the probability that two text positions decoded in parallel commit the same token,
and \emph{entropy-matched guidance} that rescales the CFG target so that its entropy matches that of the conditional prediction.
Figure~\ref{fig:teaser} shows the overall achievements of Omni-Diffusion-Distill, and Figure~\ref{fig:pipeline} gives an overview of the framework. We summarize our contributions as follows:
\begin{itemize}[leftmargin=*]
\vspace{-1em}
\setlength\itemsep{0em}
\item \textbf{Image generation task.} At $8$ decoding steps, Omni-Diffusion-Distill reaches overall scores of
$0.828$ on GenEval and $83.0$ on DPG-Bench, ahead of every distilled baseline on all five text-to-image benchmarks
and within $0.8$ points of the full-step teacher on DPG-Bench. On Graph-200K it reaches a DINOv2 image
similarity of $0.789$ against the full-step teacher's $0.799$.

\item \textbf{Multimodal understanding task.} At $64$ decoding steps, Omni-Diffusion-Distill reaches GPT judge scores
of $20.0$ on MM-Vet and $57.2$ on COCO captioning, twice the same-step teacher's $28.4$. It leads every distilled baseline
on all five open-ended benchmarks.

\item \textbf{Efficiency.} The student generates an image in $8$ steps and answers a question in $64$. On one A100 this is $18.2\times$ faster than the teacher for text-to-image
generation and $21.2\times$ for understanding.
\end{itemize}

\section{Related Work}
\label{sec:related}

Unified models differ in how images are represented: a shared discrete vocabulary
\citep{chameleon2024mixed, wang2024emu3}, a separate tokenizer for generation \citep{wu2024janus}, or
continuous latents generated with diffusion \citep{zhou2024transfusion, deng2025bagel}. Closest to us are those that
generate with an iterative discrete denoiser over a shared text and image vocabulary
\citep{yang2025mmada, xin2025luminadimoo, swerdlow2025unidisc, shi2025muddit}. Few-step
distillation methods are largely modality-specific: trajectory-based distillation and distribution matching for
image diffusion \citep{salimans2022progressive, song2023consistency, liu2023flow, yin2024dmd,
yin2024dmd2, zhou2024score}, output re-masking for discrete image generators \citep{zhu2025dimo},
objectives that train on student-generated sequences \citep{agarwal2024gkd, ko2024distillm} and
preference objectives \citep{rafailov2023dpo} for autoregressive LLMs, and, for text dLLMs, trajectory-based distillation
\citep{kim2025cdlm, deschenaux2024beyond, zhang2026t3d} and distribution matching
\citep{zheng2025ultrafast}. Unified models have been accelerated by consistency distillation
\citep{xu2025unicms} and by speculative decoding with a distilled continuous image branch
\citep{lu2025hyperbagel}. Omni-Diffusion-Distill instead performs on-policy refinement on states the student itself
produces, and treats both modalities as masked tokens of one model, so one framework serves both.
Appendix~\ref{sec:appendix-related} expands this discussion.

\section{Method}
\label{sec:method}

\subsection{Background and Preliminaries}
\label{sec:formulation}

\noindent\textbf{Why a fully discrete unified model.}
Omni-Diffusion-Distill leverages
a single
denoiser that predicts a categorical distribution at every position, for both text and image
tokens. At every step, the teacher therefore gives an explicit prediction, and
one distillation framework can learn from it directly. Most unified models do not have such a
denoiser.
Some generate images with a continuous diffusion branch or decoder \citep{you2026lladao,
li2025dualdiffusion, inclusionai2026llada2uni}. Some use separate parameters for the two modalities
\citep{li2025lavidao}. Others replace masked diffusion with discrete flow matching
\citep{wang2025fudoki, luo2025nextomni}. Only a few unified models are fully discrete
\citep{yang2025mmada, shi2025muddit, swerdlow2025unidisc, xin2025luminadimoo}.
Our method is not designed around a specific architecture. It only requires a shared discrete token space, iterative
masked decoding and classifier-free guidance for images.

\noindent\textbf{Notation.} Text tokens, vector-quantized image tokens and a $\mathsf{mask}$ token share one vocabulary. Masked diffusion learns to recover masked output tokens. Decoding starts from a fully masked output. At each step, the model commits some masked positions, replacing their $\mathsf{mask}$ tokens with sampled tokens that stay fixed. We call the partially decoded output at decoding step $t$ the state $x_t$, and write $\mathcal{M}(x_t)$ for its masked positions. At each masked position $i$, the frozen teacher $p_\phi$ and the student $q_\theta$ predict
$p_\phi^{\,i}(\cdot\mid x_t)$ and $q_\theta^{\,i}(\cdot\mid x_t)$.

\noindent\textbf{Block-causal attention.} The teacher attends bidirectionally over the whole sequence. The student uses a block-causal mask for text outputs and bidirectional attention for images.
Every position can attend to the prompt.
Text outputs are split into blocks, and each text token attends to its own block and to all earlier
blocks. At inference, the student decodes all image tokens in parallel and commits the most
confident positions at each step. Text is decoded one block at a time, so the KV cache of finished
blocks can be reused exactly.

\subsection{Motivation for Two-Stage Training}
\label{sec:overview}
\textbf{Stage 1 teaches the student to decode in fewer steps.}
The student is initialized from the teacher, but it was never trained to decode in so few steps.
With fewer steps, it must commit more tokens in each forward pass. Stage~1 teaches this offline
from teacher trajectories cached in advance. After Stage~1, the student can decode in fewer steps, but its quality is still below the teacher's (Table~\ref{tab:abl-stage}).

\textbf{Stage 2 reduces the remaining quality gap.}
Stage~1 trains only on states from the teacher's trajectories. At inference, however, the student
continues from states produced by its own earlier steps, so its errors can accumulate. Stage~2
addresses this by training the student on intermediate states of its own rollouts.
This stage also addresses the two challenges of repeated text and entropy collapse.

\subsection{Stage 1: Off-Policy Trajectory Replay for Few-Step Decoding}
\label{sec:stage1}

\noindent\textbf{Cached teacher trajectory.} For each training example, we run the teacher once
offline and cache its trajectory. The cache records the step at which each token is committed and
a few keyframes along the trajectory. Each keyframe stores the state at that step and the teacher's
predicted distributions at the positions committed since the previous keyframe. On image tasks,
we merge guidance into the teacher logits before storing them,
$\ell^{\mathrm{cfg}}=\ell^{\mathrm{uncond}}+w_{\mathrm{task}}\,(\ell^{\mathrm{cond}}-\ell^{\mathrm{uncond}})$.
Here $\ell^{\mathrm{cond}}$ and $\ell^{\mathrm{uncond}}$ are the teacher's logits at a masked
position with and without the prompt, and $w_{\mathrm{task}}$ is the guidance strength of the task.
The student therefore learns the guided distribution directly.

\noindent\textbf{Objective.} Stage~1 optimizes the same four terms for every task,
\begingroup
\setlength{\abovedisplayskip}{4pt}
\setlength{\belowdisplayskip}{4pt}
\setlength{\abovedisplayshortskip}{4pt}
\setlength{\belowdisplayshortskip}{4pt}
\begin{equation}\small
\mathcal{L}_1=\lambda_{\mathrm{traj}}\,\mathcal{L}_{\mathrm{traj}}+\lambda_{\mathrm{sc}}\,\mathcal{L}_{\mathrm{sc}}
+\lambda_{\mathrm{path}}\,\mathcal{L}_{\mathrm{path}}+\lambda_{\mathrm{gt}}\,\mathcal{L}_{\mathrm{gt}},
\label{eq:stage1}
\end{equation}
\endgroup
where the $\lambda$ are scalar weights. Each training step samples two keyframes of a cached trajectory. The trajectory term $\mathcal{L}_{\mathrm{traj}}$ is the forward KL divergence between the cached teacher predictions and the student's predictions on the earlier keyframe, at the positions the teacher commits between the two keyframes. Since the teacher made these predictions at later steps, the student learns to commit in one step what the teacher commits over several. The self-consistency term $\mathcal{L}_{\mathrm{sc}}$ uses the student's prediction at the later keyframe
as a fixed target for its prediction at the earlier one, at the positions still masked at the later keyframe. This keeps the student consistent when it
skips ahead along the trajectory. The path term $\mathcal{L}_{\mathrm{path}}$ rebuilds an
intermediate state at a random step from the cached commit step of each token and trains the student to predict the tokens the
teacher finally commits at its masked positions. The ground-truth anchor $\mathcal{L}_{\mathrm{gt}}$ is the masked
cross-entropy on the training data. Eqs.~\ref{eq:stage1-terms} and~\ref{eq:stage1-rest} in
Appendix~\ref{sec:appendix-loss} give all four terms in full, and Algorithm~\ref{alg:stage1} gives
the training loop.

\begin{wrapfigure}{r}{0.5\textwidth}
\vspace{-2.5em}
\begin{minipage}{0.5\textwidth}
\begin{algorithm}[H]
\caption{Stage 2: On-Policy Refinement}
\label{alg:stage2}
\small
\begin{algorithmic}[1]
\State \textbf{Input:} teacher $p_\phi$, student
  $q_\theta$, auxiliary network $r_\psi\!\leftarrow\!p_\phi$,
  reference $q_{\mathrm{ref}}\!\leftarrow\!q_\theta$, weights $\lambda$, training steps $N_2$,
  refresh interval $\tau$, stop-gradient $\mathrm{sg}$
\State \textbf{Output:} few-step student $q_\theta$
\For{$N_2$ training steps}
  \State sample a task; roll out $q_\theta$ \Comment{no grad}\label{ln:roll}
  \If{image task}
     \State $z_\theta\gets$ logits of $q_\theta$ at a random rollout state $x_t$
      \Comment{grad}\label{ln:img-a}
    \State $\hat{x}_0\sim\mathrm{softmax}(\mathrm{sg}[z_\theta])$;\;
      $s\gets\mathrm{remask}(\hat{x}_0)$\label{ln:remask}
    \State $\mathcal{M}\gets$ positions masked in both $s$ and $x_t$\label{ln:img-m}
    \State $p_\phi\gets$ entropy-matched teacher on $s$
      \Comment{Eq.~\ref{eq:em}}\label{ln:em}
    \State $\nabla_{\!z_\theta}\mathcal{L}^{\,\mathrm{img}}_{\mathrm{match}}
      \gets\mathrm{sg}[r_\psi-p_\phi]$\label{ln:img-g}
    \State $\mathcal{L}_{\mathrm{col}}\gets 0$\label{ln:img-b}
  \Else
    \State $s\gets$ rollout state at a random decoding step of a
      block\label{ln:txt-a}
    \State $\mathcal{M}\gets$ uncommitted positions of $s$\label{ln:txt-m}
    \State $p_\phi^{\mathrm{seq}}\gets p_\phi$ decodes $\mathcal{M}$ one at a time\label{ln:seq}
    \State $\mathcal{L}^{\,\mathrm{text}}_{\mathrm{match}}\gets
      \mathrm{JSD}(p_\phi^{\mathrm{seq}},q_\theta)$
      \Comment{grad}\label{ln:txt-jsd}
    \State $\mathcal{L}_{\mathrm{col}}\gets$ collision penalty on $\mathcal{P}$
      \Comment{Eq.~\ref{eq:rel}}\label{ln:txt-b}
  \EndIf
  \State $\mathcal{L}_{\mathrm{pref}}\gets$ preference term
    \Comment{Eq.~\ref{eq:pref}}\label{ln:pref}
  \State $\mathcal{L}_{\mathrm{gt}}\gets\mathrm{CE}$ on freshly masked data
  \State $\theta\gets\mathrm{update}(\theta,\mathcal{L}_2)$
    \Comment{$\lambda$-weighted sum}\label{ln:update}
  \State $\psi\gets\mathrm{update}(\psi,\mathrm{CE}(r_\psi,\hat{x}_0))$ on
    image tasks
  \State every $\tau$ steps: $q_{\mathrm{ref}}\gets q_\theta$
\EndFor
\end{algorithmic}
\end{algorithm}
\end{minipage}
\vspace{-3.4em}
\end{wrapfigure}
\subsection{Stage 2: On-Policy Refinement on Student Rollouts}
\label{sec:stage2}\label{sec:unified}

Stage~2 starts from the Stage-1 student (Algorithm~\ref{alg:stage2}).
Each training step samples one task and rolls out the student without gradient (line~\ref{ln:roll}). We build a state $s$ from the rollout, run the teacher on $s$, and train the student to match the teacher prediction at the scored positions $\mathcal{M}$. Both modalities share this loop and differ in the state, the teacher prediction and the matching term, so $\mathcal{L}_{\mathrm{match}}=\E_{\mathrm{task}}[\mathcal{L}^{\,\mathrm{task}}_{\mathrm{match}}]$.

\subsubsection{Image Branch: Matching the Distribution}
\label{sec:stage2-image}
This branch trains the student to match the teacher's image distribution
(lines~\ref{ln:img-a}--\ref{ln:img-b}). The student first predicts on a rollout state $x_t$
(line~\ref{ln:img-a}). We sample a clean image $\hat{x}_0$ from this prediction and re-mask it to
obtain the scored state $s$ (line~\ref{ln:remask}). The visible tokens of $s$ are thus generated by
the student. The student's prediction on $s$ does not capture the distribution of images it generates, so following distribution matching for diffusion models~\citep{yin2024dmd, yin2024dmd2, zhu2025dimo}, we estimate this distribution with an auxiliary network $r_\psi$. 
The student is updated only at the scored positions $\mathcal{M}$,
which are masked in both $s$ and $x_t$ (line~\ref{ln:img-m}). Sampling discrete tokens is not differentiable, so we cannot backpropagate
through $\hat{x}_0$. Instead, we directly set the gradient on the student's logits $z_\theta$ to the
difference between the predictions of $r_\psi$ and the teacher on $s$, at each position in
$\mathcal{M}$ (line~\ref{ln:img-g}). Meanwhile, $r_\psi$ is trained with cross-entropy on the
student's samples, $\mathrm{CE}(r_\psi(\cdot\mid s),\hat{x}_0)$, so that it keeps up with the student.

\noindent\textbf{Entropy-matched guidance.} The teacher prediction here is the CFG target
$\mathrm{softmax}(\ell^{\mathrm{cfg}})$. It is much sharper than the conditional prediction
$\mathrm{softmax}(\ell^{\mathrm{cond}})$ at every step, and its entropy changes little along the
trajectory (Figure~\ref{fig:entropy_temperature}). A student trained to fit it keeps losing entropy over the decoding steps and ends even sharper than the target (Figure~\ref{fig:motivation} and Table~\ref{tab:analysis}).
Such a sharp student tends to leave the source image unchanged in editing.
We therefore rescale the CFG logits
with a temperature $\nu$ (line~\ref{ln:em}).
At each scored position, we choose $\nu$ so that the entropy $H$ of the rescaled target is a
fixed fraction $\rho$ of the entropy of the conditional prediction,
\begingroup
\setlength{\abovedisplayskip}{4pt}
\setlength{\belowdisplayskip}{4pt}
\setlength{\abovedisplayshortskip}{4pt}
\setlength{\belowdisplayshortskip}{4pt}
\begin{equation}\small
H\big(\mathrm{softmax}(\ell^{\mathrm{cfg}}/\nu)\big)=\rho\,H\big(\mathrm{softmax}(\ell^{\mathrm{cond}})\big),
\qquad
p_\phi \;\leftarrow\; \mathrm{softmax}\big(\ell^{\mathrm{cfg}}/\nu\big).
\label{eq:em}
\end{equation}
\endgroup
We solve for $\nu$ by bisection (Appendix~\ref{sec:appendix-loss}). The rescaled distribution $p_\phi$ is then used as the teacher prediction in line~\ref{ln:img-g}.
Temperature scaling changes only the entropy and keeps the ranking of tokens.

\subsubsection{Text Branch: Sequential Targets and Penalty on Parallel Collisions}
\label{sec:stage2-text}
The text branch differs from the image branch in two ways
(lines~\ref{ln:txt-a}--\ref{ln:txt-b}). First, it matches the teacher at each position instead of
over the whole output. The student predicts these per-position distributions directly, so no auxiliary network is
needed. Second, its main challenge is repetition under parallel decoding. 
We take
$s$ as the rollout state at a random step (line~\ref{ln:txt-a}), and $\mathcal{M}$ as the positions
not yet committed in $s$ (line~\ref{ln:txt-m}).

\noindent\textbf{Sequentially conditioned teacher predictions.} Running the teacher once on $s$
gives its one-step marginal $p_\phi^{\,i}(\cdot\mid s)$ at each position. This marginal does not
know which tokens the other positions will commit. We instead let the teacher decode the scored
positions one at a time (line~\ref{ln:seq}). At each position $i$, the teacher commits a token,
and we record the distribution it used, denoted $p_\phi^{\mathrm{seq},i}$. This
distribution is conditioned on the tokens the teacher has already committed. The student therefore
learns from sequential context, while its own decoding stays parallel. The matching term is the
Jensen-Shannon divergence between $p_\phi^{\mathrm{seq},i}$ and $q_\theta^{\,i}$, averaged over
the positions in $\mathcal{M}$ (line~\ref{ln:txt-jsd}).

\noindent\textbf{Penalty on repetitions in parallel decoding.} $\mathcal{L}^{\,\mathrm{text}}_{\mathrm{match}}$
is averaged over positions, so it is already minimal when each position matches its own teacher
prediction. However, positions committed in the same decoding step are sampled independently, and
the matching term does not penalize repeated tokens among them. When a token is plausible at two
positions of a block, the student may commit it at both. We call this chance the collision
probability. For positions $i$ and $j$, it equals the inner product
$\langle q_\theta^{\,i},q_\theta^{\,j}\rangle$. All scored positions of a block are predicted from
the same state $s$, and any two of them may be committed in the same decoding step. We therefore
penalize collisions over the set $\mathcal{P}$ of all pairs of scored positions (line~\ref{ln:txt-b}),
\begingroup
\setlength{\abovedisplayskip}{2pt}
\setlength{\belowdisplayskip}{2pt}
\setlength{\abovedisplayshortskip}{2pt}
\setlength{\belowdisplayshortskip}{2pt}
\begin{equation}\small
\mathcal{L}_{\mathrm{col}}=\frac{1}{|\mathcal{P}|}\sum_{(i,j)\in\mathcal{P}}
-\log\big(1-\langle q_\theta^{\,i},q_\theta^{\,j}\rangle\big).
\label{eq:rel}
\end{equation}
\endgroup
The penalty reduces but does not forbid repetition. A token can still appear at two positions if the student assigns it a high probability at both. We apply it only to text, where parallel decoding often causes repetition.

\noindent\textbf{Stage-2 objective.} Stage~2 optimizes
$\mathcal{L}_2=\lambda_{\mathrm{match}}\mathcal{L}_{\mathrm{match}}+\lambda_{\mathrm{col}}\mathcal{L}_{\mathrm{col}}+\lambda_{\mathrm{pref}}\mathcal{L}_{\mathrm{pref}}+\lambda_{\mathrm{gt}}\mathcal{L}_{\mathrm{gt}}$.
Here $\mathcal{L}_{\mathrm{match}}$ is the expectation over tasks defined above, $\mathcal{L}_{\mathrm{col}}$
is Eq.~\ref{eq:rel}, and $\mathcal{L}_{\mathrm{gt}}$ is the same ground-truth anchor as in Stage~1.
The preference term $\mathcal{L}_{\mathrm{pref}}$ follows T3D~\citep{zhang2026t3d} (Appendix~\ref{sec:appendix-loss}).

\begin{figure}[t]
\includegraphics[width=\linewidth]{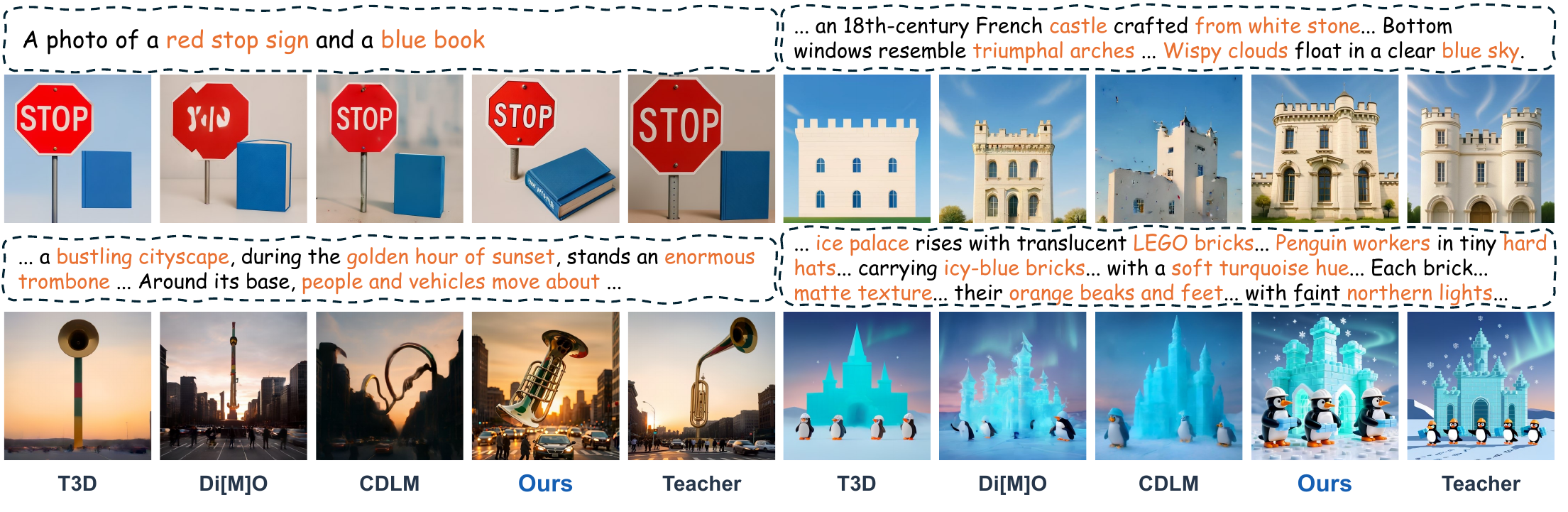}
\vspace{-2em}
\caption{\textbf{Qualitative comparison on T2I.}
All methods use the same prompts and seeds.
Distilled models use 8 decoding steps, and the teacher uses 128.}
\vspace{-1.8em}
\label{fig:qual-t2i}
\end{figure}
\section{Experiments}
\label{sec:exp}

\begin{figure}[h]
\centering
\includegraphics[width=\linewidth]{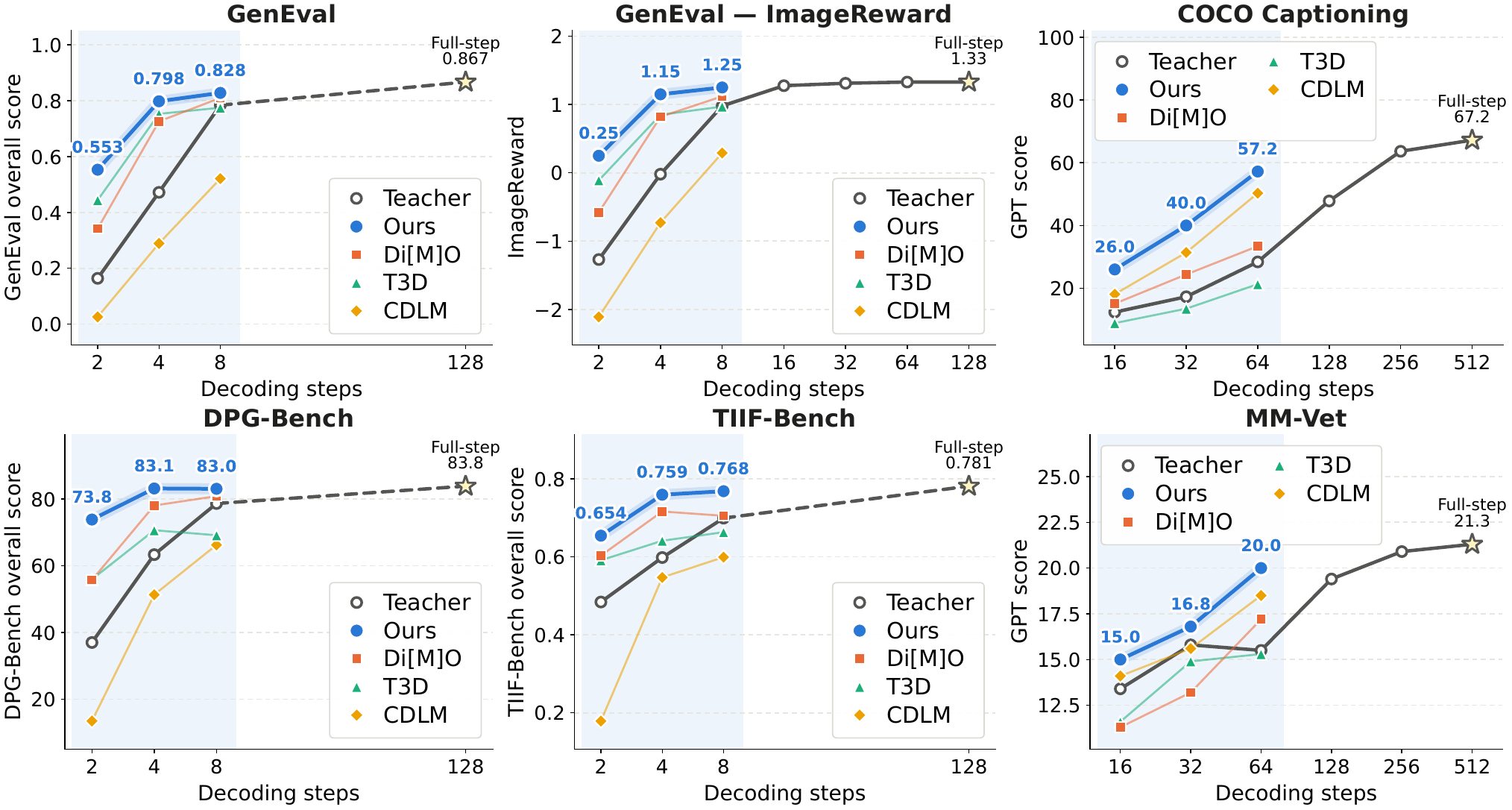}
\vspace{-2em}
\caption{\textbf{Performance across decoding budgets.} The first two columns show T2I benchmarks, and the last column shows MMU benchmarks. The teacher curve extends to the full-step budget.}
\vspace{-1.5em}
\label{fig:step-quality}
\end{figure}
\subsection{Experimental Setup}
\label{sec:impl}\label{sec:evalsetup}
\noindent\textbf{Model and data.} We distill Lumina-DiMOO~\citep{xin2025luminadimoo}, an $8$B fully discrete unified dLLM whose single network covers all four tasks. Training uses approximately $51$K examples, and Appendices~\ref{sec:appendix-data} and~\ref{sec:appendix-cost} give the data sources and training settings.

\noindent\textbf{Benchmarks and baselines.} Following the teacher's protocol for most benchmarks, we evaluate the four
tasks in three groups on 17 benchmarks: text-to-image generation (T2I), image-to-image generation (I2I) and
multimodal understanding (MMU). I2I covers image editing and controllable generation. MMU uses both
discriminative and open-ended benchmarks. We also report the teacher at both the few-step and
full-step budgets. The distilled baselines are Di[M]O~\citep{zhu2025dimo}, CDLM~\citep{kim2025cdlm} and T3D~\citep{zhang2026t3d}. Appendices~\ref{sec:appendix-baselines} and~\ref{sec:appendix-benchlist} give more details.

\noindent\textbf{Decoding budgets and cost.} Image tasks use $8$, $4$ or $2$ decoding steps,
against $128$ for the teacher. Understanding uses $64$, $32$ or $16$ steps, against $512$ for the
teacher. We report the $8$-step and $64$-step results here and the smaller budgets in
Appendix~\ref{sec:appendix-budgets}. Figure~\ref{fig:step-quality} shows how performance changes with the decoding steps on T2I and MMU benchmarks.  We also report the number of function evaluations (NFE), the decoding steps times the forward passes per step. The student uses one pass per step, while the teacher adds one pass per guidance term, giving two on T2I and three on I2I.

\subsection{T2I: Outperforming Every Distilled Baseline and the Few-Step Teacher}
\label{sec:res-t2i}

Table~\ref{tab:main-t2i} reports the overall scores on five T2I benchmarks. They cover compositional
prompts, dense prompt following, text rendering, reasoning, style and long instructions. Figure~\ref{fig:qual-t2i}
compares samples under the same prompts and seeds. Omni-Diffusion-Distill is the best distilled method on all five benchmarks. At $8$ steps, it
reaches $0.828$ on GenEval and $83.0$ on DPG-Bench. It is above the few-step teacher on every benchmark at
half its NFE, and within $0.8$ DPG-Bench points of the full-step teacher. With fewer steps, its lead over
the baselines generally becomes larger. Appendix~\ref{sec:appendix-bench} gives more results.

\begin{table}[t]
\caption{\textbf{Quantitative Results on T2I Benchmarks.} Overall score on each benchmark. ImageReward~\citep{xu2023imagereward} and Aesthetic are image quality metrics computed
on GenEval. \bestmark{} and \secondmark{} mark the top two distilled results.}
\vspace{-1em}
\label{tab:main-t2i}
\begin{center}
\setlength{\tabcolsep}{4pt}
\fit{%
\small
\begin{tabular}{lc ccc cccc}
\toprule
\multirow{2}{*}{Method} & \multirow{2}{*}{NFE $\downarrow$} & \multicolumn{3}{c}{GenEval $\uparrow$} & \multirow{2}{*}{DPG-Bench} & \multirow{2}{*}{UniGenBench} & \multirow{2}{*}{OneIG} &
\multirow{2}{*}{TIIF-Bench} \\
\cmidrule(lr){3-5}
& & Overall & ImageReward & Aesthetic & & & & \\
\midrule
Teacher (full-step)  &  $128{\times}2$  & 0.867 & 1.33 & 3.65 &  83.8  &  0.722  &  0.445  &  0.781 \\
\midrule
T3D     &  $8{\times}1$  & 0.775 & 0.97 & 3.55 &  69.1  &  0.640  &  0.306  &  0.663 \\
Di[M]O  &  $8{\times}1$  & \snd{0.810} & \snd{1.12} & \snd{3.61} &  \snd{80.8}  &  \snd{0.700}  &  \snd{0.327}  &  \snd{0.705} \\
CDLM    &  $8{\times}1$  & 0.521 & 0.29 & 3.33 &  66.2  &  0.536  &  0.280  &  0.599 \\
\textbf{Omni-Diffusion-Distill}  &  $8{\times}1$  & \bst{0.828} & \bst{1.25} & \bst{3.71} &  \bst{83.0}  &  \bst{0.710}  &  \bst{0.381}  &  \bst{0.768} \\
Teacher (few-step)  &  $8{\times}2$  & 0.784 & 0.98 & 3.61 &  78.6  &  0.676  &  0.357  &  0.699 \\
\bottomrule
\end{tabular}}
\vspace{-1.8em}
\end{center}
\end{table}
\subsection{I2I: Editing Degrades Little Under Step Reduction}
\label{sec:res-i2i}

\begin{figure}[t]
\centering
\includegraphics[width=\linewidth]{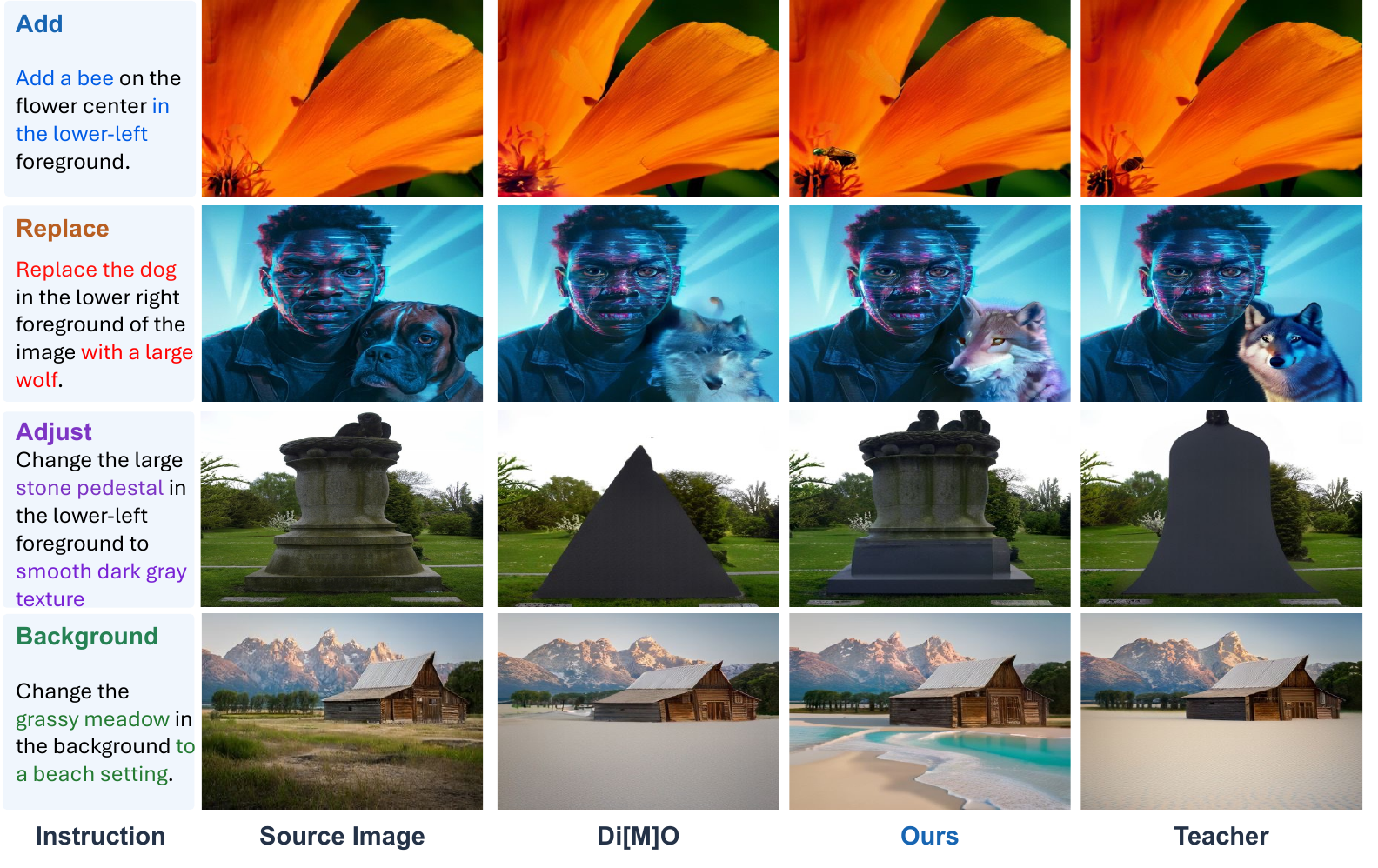}
\vspace{-2.7em}
\caption{\textbf{Qualitative comparison on image editing.}
We compare with the teacher and the strongest distilled baseline Di[M]O.
Distilled models use 8 decoding steps, and the teacher uses 128.}
\label{fig:qual-i2i}
\vspace{-0.5em}
\end{figure}

Table~\ref{tab:main-i2i} reports image editing on the I2I benchmark and controllable generation on Graph-200K.
Figure~\ref{fig:qual-i2i} shows four edit types, and Figure~\ref{fig:qual-i2i-more} shows more results. \textbf{Editing quality drops little with fewer steps.} When the teacher goes from $128$ to $8$ steps, the I2I benchmark drops by $2.6\%$ and Graph-200K by $0.9\%$, consistent with other editing models~\citep{yan2026specedit,liu2025lgcc}. The source image already determines most of the output’s layout and appearance~\citep{lu2025hyperbagel}, so tokens
outside the edited region can be committed together at little cost~\citep{chen2025regione}. This leaves distillation less room to help. 

\noindent\textbf{Entropy-matched guidance helps the student follow the instruction.}  Without it, the student's entropy keeps falling over the decoding steps and ends at $1.55$, below the $2.28$ of the teacher's CFG target (Figure~\ref{fig:motivation} and Table~\ref{tab:analysis}). Such an overconfident student tends to copy the source image instead of making the edit. With entropy-matched guidance, the final entropy stays at $3.32$, and the I2I benchmark score rises from $2.04$ to $2.31$. This $13\%$ gain is the largest in Table~\ref{tab:abl-em}.

\noindent\textbf{Omni-Diffusion-Distill leads on Add and Compose.} On Add it scores $3.26$, while the best other distilled method scores $2.52$. It ranks second on the I2I benchmark Overall and Graph-200K. It is also above the few-step teacher on Compose and Adjust, although the teacher uses $24$ NFE and the student uses $8$. On Graph-200K it scores $0.789$, while the full-step teacher scores $0.799$ with $48\times$ more forward passes. At $4$ and $2$ steps, it beats the few-step teacher on Graph-200K (Table~\ref{tab:main-i2i-more}).

\begin{table}[t]
\vspace{-1em}
\caption{\textbf{Quantitative Results on I2I. The few-step teacher is already strong, so distillation has less room to help. Omni-Diffusion-Distill is on par with the best distilled baseline.} I2I benchmark scores range from 1 to 5. Overall is the mean of nine edit types, and four are shown. Graph-200K reports DINOv2 image similarity.}
\vspace{-1em}
\label{tab:main-i2i}
\begin{center}
\fit{%
\begin{tabular}{lc ccccc c}
\toprule
\multirow{2}{*}{Method} & \multirow{2}{*}{NFE $\downarrow$} & \multicolumn{5}{c}{\textbf{I2I Benchmark}} & \textbf{Graph-200K} \\
\cmidrule(lr){3-7} \cmidrule(lr){8-8}
& & Add $\uparrow$ & Compose $\uparrow$ & Adjust $\uparrow$ & Style $\uparrow$ & Overall $\uparrow$ & DINOv2 $\uparrow$ \\
\midrule
Teacher (full-step) & $128{\times}3$ & 3.62 & 1.94 & 2.29 & 3.89 & 2.66 & 0.799 \\
\midrule
T3D & $8{\times}1$ & 2.12 & 1.78 & \bst{2.49} & 3.72 & 2.28 & 0.774 \\
Di[M]O & $8{\times}1$ & \snd{2.52} & \snd{1.85} & 2.16 & \bst{3.90} & \bst{2.38} & \bst{0.794} \\
CDLM & $8{\times}1$ & 1.47 & 1.51 & 1.45 & 3.11 & 1.68 & 0.657 \\
\textbf{Omni-Diffusion-Distill} & $8{\times}1$ & \bst{3.26} & \bst{1.90} & \snd{2.45} & \snd{3.83} & \snd{2.31} & \snd{0.789} \\
Teacher (few-step) & $8{\times}3$ & 3.32 & 1.56 & 2.41 & 3.93 & 2.59 & 0.792 \\
\bottomrule
\end{tabular}}
\vspace{-2em}
\end{center}
\end{table}

\subsection{MMU: Step Reduction Primarily Affects Open-Ended Generation}
\label{sec:res-mmu}
\begin{figure}[t]
\centering
\includegraphics[width=\linewidth]{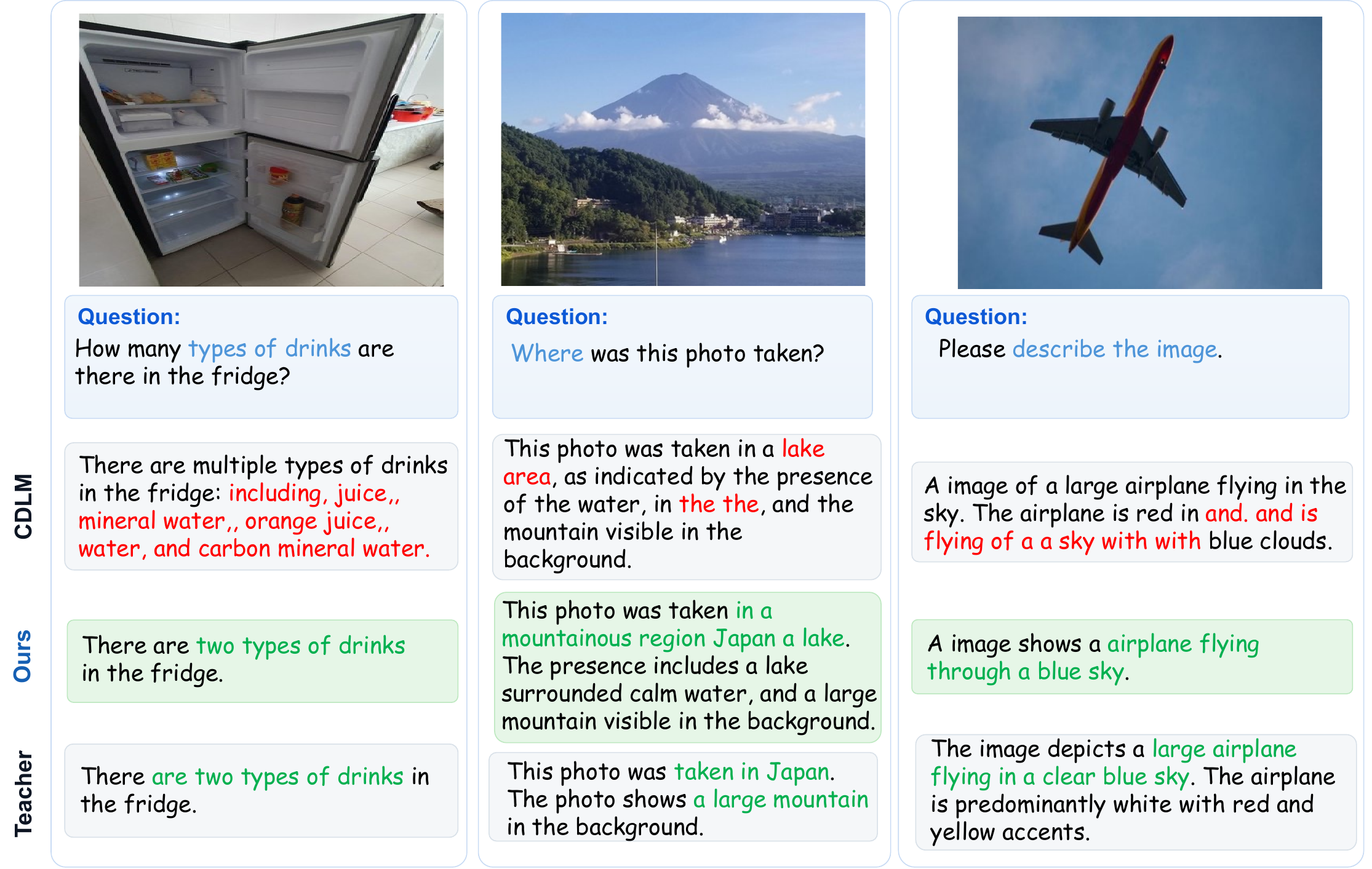}
\vspace{-1.5em}
\caption{\textbf{Qualitative comparison on MMU.} We compare with the teacher and the strongest distilled baseline CDLM. Distilled models use 64 decoding steps, and the teacher uses 512.  \textcolor{red}{Red} marks repeated or incorrect text, and \textcolor[HTML]{2BA84A}{green} marks semantically correct content.}
\label{fig:qual-mmu}
\vspace{-1.5em}
\end{figure}

\begin{table}[h]
\caption{\textbf{Quantitative Results on Discriminative MMU. Discriminative scores remain stable with fewer steps.} All benchmarks use their standard parser and metric.}
\label{tab:mmu-disc}
\begin{center}
\small
\fit{
\begin{tabular}{lcccccc}
\toprule
Method & NFE $\downarrow$ & POPE $\uparrow$ & MME-P $\uparrow$ & MMBench $\uparrow$ & SEED $\uparrow$ & MMMU $\uparrow$ \\
\midrule
Teacher (full-step) & $512$ & 86.8 & 1552 & 91.2 & 85.3 & 58.3 \\
\midrule
T3D & $64$ & 82.8 & 1429 & 60.2 & 78.6 & 37.8 \\
Di[M]O & $64$ & \snd{87.0} & \snd{1458} & 57.0 & \snd{79.5} & 41.2 \\
CDLM & $64$ & 79.5 & 1344 & \snd{78.4} & 79.2 & \snd{50.7} \\
\textbf{Omni-Diffusion-Distill} & $64$ & \bst{87.4} & \bst{1537} & \bst{89.4} & \bst{84.6} & \bst{56.3} \\
Teacher (few-step) & $64$ & 86.4 & 1552 & 90.8 & 85.1 & 58.2 \\
\bottomrule
\end{tabular}}
\vspace{-1.2em}
\end{center}
\end{table}

\begin{table}[!t]
\caption{\textbf{Quantitative Results on Open-Ended MMU.} A GPT judge scores the outputs. LLaVA-Bench reports a relative score, and DeCapBench reports DCSCORE.}
\vspace{-1.2em}
\label{tab:mmu-open}
\begin{center}
\small
\fit{
\begin{tabular}{lc ccccc}
\toprule
Method & NFE $\downarrow$ & MM-Vet $\uparrow$ & Vibe-Eval $\uparrow$ & LLaVA-Bench $\uparrow$ & COCO $\uparrow$ & DeCapBench $\uparrow$ \\
\midrule
Teacher (full-step) & $512$ & 21.3 & 12.36 & 25.1 & 67.2 & 44.5 \\
\midrule
T3D & $64$ & 15.3 & 3.81 & 14.7 & 21.3 & 13.8 \\
Di[M]O & $64$ & 17.2 & 6.04 & 14.5 & 33.4 & 15.9 \\
CDLM & $64$ & \snd{18.5} & \snd{7.71} & \snd{15.6} & \snd{50.3} & \snd{16.9} \\
\textbf{Omni-Diffusion-Distill} & $64$ & \bst{20.0} & \bst{8.18} & \bst{16.9} & \bst{57.2} & \bst{19.6} \\
Teacher (few-step) & $64$ & 15.5 & 5.30 & 14.7 & 28.4 & 16.2 \\
\bottomrule
\end{tabular}}
\vspace{-2.7em}
\end{center}
\end{table}

We evaluate MMU with two groups of benchmarks. Discriminative benchmarks check a short answer, and open-ended benchmarks use a GPT judge on the whole output. Tables~\ref{tab:mmu-disc} and~\ref{tab:mmu-open} report the two
groups, and Appendix~\ref{sec:appendix-judge} gives the judging protocols.

\noindent\textbf{Discriminative scores remain stable with fewer steps.} When the teacher goes
from $512$ to $64$ steps, all five scores stay within half a point. MME-P stays at $1552$, and MMMU
moves from $58.3$ to $58.2$. Omni-Diffusion-Distill is just as stable from $64$ down to $16$ steps.
It is also the best distilled method on all five benchmarks.

\noindent\textbf{Open-ended scores drop sharply with fewer steps, and Omni-Diffusion-Distill
recovers much of this loss.} These benchmarks check the whole answer, so repeated words and broken
sentences lower the score. At $64$ steps, the teacher loses more than half of its COCO captioning score and
about two thirds of its DeCapBench score. Omni-Diffusion-Distill beats the few-step teacher on all
five open-ended benchmarks. On COCO captioning it scores $57.2$, while the few-step teacher scores $28.4$.
In Figure~\ref{fig:qual-mmu}, CDLM repeats words and leaves
phrases unfinished, and Table~\ref{tab:analysis} measures this repetition. A gap to the
full-step teacher remains, and it is largest on DeCapBench. Appendices~\ref{sec:appendix-results}
and~\ref{sec:appendix-eval} give more results.

\section{Conclusion}
\label{sec:conclusion}
We present Omni-Diffusion-Distill, a few-step distillation framework for fully discrete unified
multimodal dLLMs. The student learns to skip decoding steps from cached teacher trajectories and
is then refined on its own rollouts, with one on-policy loop for images and text. To address the
challenge each modality faces, entropy-matched guidance keeps the image student from becoming
overconfident, and a pairwise collision penalty reduces repetition in parallel text decoding.
Distillation also makes the text attention block-causal, so understanding reuses its KV cache exactly. The student generates an image in $8$ steps and answers a question in $64$, $18.2\times$ and
$21.2\times$ faster than the teacher, and outperforms all distilled baselines on most benchmarks, even approaching the full-step teacher on several.
\label{sec:limits}
Our approach is limited to unified dLLMs that represent text and images as discrete tokens, and we
validate it only on Lumina-DiMOO. The student still trails the full-step teacher, most on text
rendering and detailed captioning. On image editing, it also trails the few-step teacher at $8$ steps, although it uses a
third of the NFE.

\bibliography{references}
\bibliographystyle{iclr2027_conference}

\newpage
\appendix
\section{Extended Related Work}
\label{sec:appendix-related}

\subsection{Unified Multimodal Understanding and Generation}
Unified models differ mainly in how they represent and generate images alongside text.
Autoregressive models generate image tokens with the same transformer that generates text
\citep{chameleon2024mixed, wang2024emu3, lu2023unifiedio2}. Janus and Janus-Pro keep this design
but use separate visual encoders for understanding and generation \citep{wu2024janus,
chen2025januspro}. Hybrid designs keep text autoregressive but generate images with continuous
diffusion, either inside the same transformer or through a separate decoder or branch
\citep{zhou2024transfusion, sun2023emu, deng2025bagel, wu2025omnigen2}.

A third family generates with an iterative discrete denoiser. One group uses masked diffusion over a
shared text and image vocabulary for both understanding and generation. It includes MMaDA and its
parallel variant \citep{yang2025mmada, tian2025mmadaparallel}, Lumina-DiMOO
\citep{xin2025luminadimoo}, UniDisc \citep{swerdlow2025unidisc}, Muddit \citep{shi2025muddit} and
Omni-Diffusion \citep{li2026omnidiffusion}. Lavida-O also denoises VQ image tokens with masked
diffusion, but processes them with a separate and smaller generation branch \citep{li2025lavidao}.
Unified LLaDA variants keep a discrete text denoiser but attach a continuous image decoder
\citep{inclusionai2026llada2uni, you2026lladao}. A closely related group replaces masked diffusion
with discrete flow matching and keeps the same iterative denoising formulation
\citep{wang2025fudoki, luo2025nextomni, susladkar2026unidflow}. D-DiT combines discrete text
diffusion with continuous image diffusion in one bidirectional model \citep{li2025dualdiffusion}.
A separate line uses dLLM decoding only for understanding. These models pair a masked diffusion text
decoder with a vision encoder and do not generate images \citep{you2025lladav, li2025lavida,
yu2025dimple}.

\subsection{Diffusion Large Language Models and Discrete Generative Modeling}
Discrete diffusion extends denoising to categorical data. Early work defines the forward corruption
as a discrete Markov process with uniform, absorbing or structured kernels
\citep{austin2021d3pm, hoogeboom2021argmax}, and continuous-time formulations give tractable
objectives and samplers \citep{campbell2022ctmc}. The absorbing kernel, also called the mask kernel,
works best for language. Score entropy \citep{lou2024sedd}, simplified masked diffusion ELBOs
\citep{sahoo2024mdlm, shi2024md4} and reparameterizations that view masked diffusion as modeling
clean-data conditionals \citep{ou2025radd} steadily closed the gap to autoregressive models. This
line led to dLLMs at LLM scale \citep{nie2025llada, ye2025dream}. Another thread interpolates between
autoregression and diffusion, with block-wise \citep{arriola2025blockdiffusion}, semi-autoregressive
\citep{han2023ssdlm} or position-dependent \citep{wu2023ardiffusion} decoding, or with continuous
diffusion over embeddings \citep{li2022diffusionlm, gong2023diffuseq, lin2023genie}. A key property
of dLLMs is that they predict an explicit distribution at each position. In continuous diffusion, the
distribution of the generator is implicit.

\subsection{Distillation for Few-Step Generation}
\label{sec:related-distill}
Few-step distillation has developed along three largely separate lines, for image generators,
autoregressive LLMs and dLLMs. We review them in turn, and then discuss unified models and on-policy distillation. Table~\ref{tab:distill} compares
representative methods by whether training is on-policy, whether the target is a fixed teacher or a
snapshot of the student, whether the generator is discrete, and the modality.

\begin{table}[t]
\caption{\textbf{Representative Methods for Few-Step and Accelerated Generation.} Methods are
grouped by model class. Policy: whether the student is trained on its own generations. Target/ref.:
T is a fixed teacher, and S is a snapshot or self-reference of the student. Disc.: discrete-token
generator. Modality: I is image, T is text, and M is unified multimodal. A dash marks methods that
are not distillation objectives, so policy and target do not apply.}
\label{tab:distill}
\begin{center}
\fit{%
\begin{tabular}{llcccc}
\toprule
Method & Family & Policy & Target/ref. & Disc. & Modality \\
\midrule
\multicolumn{6}{l}{Image generators} \\
Progressive distillation \citep{salimans2022progressive} & trajectory & off & T (iterated) & & I \\
Consistency models \citep{song2023consistency} & trajectory & off & T+S & & I \\
DMD \citep{yin2024dmd} & dist.-match & on & T (+auxiliary network) & & I \\
DMD2 \citep{yin2024dmd2} & dist.-match + adv. & on & T (+auxiliary network) & & I \\
DiffusionOPD \citep{li2026diffusionopd} & on-policy KD & on & T & & I \\
Di[M]O \citep{zhu2025dimo} & dist.-match & on & T (+auxiliary network) & \checkmark & I \\
Di4C \citep{hayakawa2025distillation} & trajectory & off & T+S & \checkmark & I, T \\
\midrule
\multicolumn{6}{l}{Autoregressive LLMs} \\
GKD \citep{agarwal2024gkd} & on-policy KD & on & T & \checkmark & T \\
\midrule
\multicolumn{6}{l}{dLLMs} \\
CDLM \citep{kim2025cdlm} & trajectory & off & T+S & \checkmark & T \\
SDTT \citep{deschenaux2024beyond} & trajectory & off & T (round-refreshed) & \checkmark & T \\
T3D \citep{zhang2026t3d} & trajectory + pref. & off & T + S (round-refreshed) & \checkmark & T \\
DiDi-Instruct \citep{zheng2025ultrafast} & dist.-match + adv. & on & T (+auxiliary network) & \checkmark & T \\
AR$\to$Diff.\ OPD \citep{su2026dataefficient} & on-policy KD & on & T & \checkmark & T \\
D2F \citep{wang2025d2f} & block-AR distill. & off & T & \checkmark & T \\
Infinite-mask \citep{yoo2026infinitemask} & mask parameterization & --- & --- & \checkmark & T \\
Fast-dLLM \citep{wu2025fastdllm} & training-free cache + parallel decoding & --- & --- & \checkmark & T \\
\midrule
\multicolumn{6}{l}{Unified multimodal (understanding + generation)} \\
UniCMs \citep{xu2025unicms} & trajectory & off & T+S & \checkmark & M \\
Hyper-Bagel \citep{lu2025hyperbagel} & multi-stage & mixed & T+S (+auxiliary network) & & M \\
\textbf{Omni-Diffusion-Distill} & \textbf{dist.-match + on-policy KD + pref.} & \textbf{off$\to$on} & \textbf{T + S (refreshed)} & \checkmark & \textbf{M} \\
\bottomrule
\end{tabular}%
}
\end{center}
\end{table}
\paragraph{Distilling image generators.}
Few-step distillation of continuous image diffusion models is well studied. Trajectory-based
methods distill several teacher steps into one. Examples include progressive distillation
\citep{salimans2022progressive}, guidance distillation \citep{meng2023distillation}, consistency and
trajectory models \citep{song2023consistency, kim2024consistency, luo2023latent}, and rectified flow
\citep{liu2023flow, liu2024instaflow}. Distribution matching and score-based methods instead match
the output distributions of the student and the teacher. They usually need a learned auxiliary
network, because the score of the continuous student cannot be evaluated directly
\citep{yin2024dmd, yin2024dmd2, zhou2024score}. Uni-DAD combines distribution matching distillation
with few-shot domain adaptation in a single stage \citep{bahram2026unidad}. Adversarial variants add a discriminator to improve
sharpness at one to four steps \citep{sauer2023adversarial, sauer2024fast}. On-policy variants train the student
on its own generations to remove the mismatch between training and inference
\citep{yin2024dmd2, kohler2024imagine, huang2025self, cui2025selfforcing}. Distillation for discrete
image generators, which are masked or autoregressive models over VQ codes, is less studied but
growing. Di[M]O distills a masked image generator into a one-step sampler with token-level
distribution matching on re-masked outputs. It keeps entropy in the student by randomizing the token
initialization \citep{zhu2025dimo}, and its soft-embedding variant further improves one-step quality
\citep{zhu2025softdimo}. Other work distills discrete diffusion image generators through
dimensional correlations \citep{hayakawa2025distillation} or fixed-point iteration
\citep{wang2026fixedpoint}. For autoregressive image models, one-step samplers have
been obtained with flow matching \citep{liu2024distilleddecoding}, and knowledge distillation has
been applied to visual autoregressive models \citep{peruzzo2026kdvar}. Apart from distillation,
other methods speed up masked and autoregressive image sampling with better
unmasking schedules \citep{besnier2025halton, you2025emigm}, speculative Jacobi decoding
\citep{teng2025sjd} or reward-tuned step reduction \citep{subbaraman2025speedrl}.

\paragraph{Distilling language models.}
For autoregressive LLMs, classical knowledge distillation matches teacher logits
\citep{hinton2015distilling}. Off-policy matching suffers from exposure bias, because the student's
own generations drift away from the teacher's. On-policy methods address this by training on
sequences generated by the student, as in generalized KD, which supports divergences such as the
generalized Jensen-Shannon divergence \citep{agarwal2024gkd}. DistiLLM combines a skew KL objective
with an adaptive off-policy scheme that reuses student-generated sequences \citep{ko2024distillm}.
Another line uses preference objectives such as direct preference optimization
\citep{rafailov2023dpo}, which pushes the model toward preferred responses and away from
dispreferred ones. Direct discriminative optimization carries this parameterization to visual
generative models and contrasts real data with samples from a frozen reference model
\citep{zheng2025ddo}.

\paragraph{Distilling diffusion large language models.}
Distillation of dLLMs is recent. For text, consistency and trajectory-based methods distill dLLMs
to a few steps \citep{kim2025cdlm, deschenaux2024beyond, zhang2026t3d, zhou2026tad}, and a discrete
distribution matching objective adapts Diff-Instruct to dLLMs \citep{zheng2025ultrafast}. Few-step
dLLMs also make errors because they commit positions independently. Infinite-state masking reduces
this error \citep{yoo2026infinitemask}, confidence thresholds address it at inference
\citep{wu2025fastdllm}, and dimensional-correlation distillation targets it during training
\citep{hayakawa2025distillation}. Many other works accelerate dLLM inference without full
distillation. Training-free methods use KV caching and parallel decoding \citep{wu2025fastdllm}, or
adaptive sampling and cache eviction \citep{wei2025slowfast, li2025prophet, liu2025dllmcache,
song2025sparsedllm}. Trained methods shorten the schedule through certainty forcing, block diffusion
fine-tuning, or distillation into a block-autoregressive hybrid \citep{chen2025dparallel,
wu2025fastdllmv2, wang2025d2f}. Each of these methods targets a single modality.

\paragraph{Distilling unified multimodal models.}
Closest to us are the few works that distill or accelerate a unified model for understanding and
generation. UniCMs \citep{xu2025unicms} applies consistency distillation to Show-o on both
modalities, but its text side follows an autoregressive parallel decoding trace instead of a
diffusion process. Hyper-Bagel \citep{lu2025hyperbagel} accelerates BAGEL with speculative decoding
and a multi-stage image pipeline. The first stage distills classifier-free guidance, and the image
generator is a continuous diffusion model. Some unified dLLMs build few-step distillation into their
generation decoder \citep{inclusionai2026llada2uni}. Others distill a unified teacher into an
understanding-only student \citep{li2026visualopsd}.

\paragraph{On-policy distillation.}
On-policy distillation (OPD) trains the student on its own generations instead of teacher-forced
data \citep{agarwal2024gkd, song2026survey, zhang2026formula}. It has developed rapidly for language
models. Recent work improves stability by mixing forward and reverse KL according to teacher entropy
\citep{jin2026entropy}, analyzes when and why OPD helps \citep{li2026rethinking}, and extends OPD to
dLLMs \citep{su2026dataefficient}. In the last work, a block-causal student is trained on its own
rollouts and supervised by the frozen autoregressive model it was converted from. Trajectory-based
distillation for dLLMs instead supervises the student on states from the teacher's own rollouts
\citep{zhang2026t3d}. Another thread makes the student its own teacher through a self-reference
\citep{zhao2026selfdistilled}. \citet{li2026diffusionopd} distill a student along its own rollouts
with a closed-form KL at each step of the sampler trajectory. They cover several continuous image
diffusion tasks, each with its own teacher. Existing OPD methods each handle one modality.

Omni-Diffusion-Distill differs from these works in two ways. First, both modalities are masked tokens of one model. A single on-policy framework therefore covers both, with a modality-specific state, teacher prediction and matching term.
Second, a pairwise collision penalty for parallel text decoding and entropy-matched guidance for
images address the degradation that step reduction causes in each modality.

\section{Algorithms and Loss Details}
\label{sec:appendix-alg}
\label{sec:appendix-loss}

\begin{algorithm}[h]
\caption{Offline Teacher Trajectory Collection}
\label{alg:rollout}
\begin{algorithmic}[1]
\State \textbf{Input:} dataset $\mathcal{O}$ of prompt-output pairs, teacher $p_\phi$, number of teacher decoding steps $T$, keyframe steps $\mathcal{F}\subseteq\{1,\dots,T\}$, number of kept probabilities $R$
\State \textbf{Output:} cache $\mathcal{D}$. Each entry stores the prompt and output, the decoding step at which each output position is committed, and one record per keyframe. Each record holds the keyframe state, together with the top-$R$
  teacher log-probabilities and the residual probability at the positions committed since the previous keyframe
\State $\mathcal{D}\gets\emptyset$
\For{each pair in $\mathcal{O}$}
  \State mask all output positions
  \For{$t=1,\dots,T$}
    \State compute the teacher prediction at the masked positions, merged with CFG on image tasks
    \State commit the positions chosen by the teacher's schedule at this step
    \State record their sparse teacher predictions and the current step
    \If{$t\in\mathcal{F}$}
      \State store a keyframe from the current state and the records accumulated since the previous one
    \EndIf
  \EndFor
  \State append the example to $\mathcal{D}$
\EndFor
\end{algorithmic}
\end{algorithm}

Algorithm~\ref{alg:rollout} collects the teacher trajectories that Stage~1 replays.
Algorithm~\ref{alg:stage1} gives Stage~1, and Algorithm~\ref{alg:stage2} in the main text gives Stage~2.

\begin{algorithm}[t]
\caption{Stage 1: Off-Policy Trajectory Replay}
\label{alg:stage1}
\begin{algorithmic}[1]
\State \textbf{Input:} cache $\mathcal{D}$ (Algorithm~\ref{alg:rollout}), teacher $p_\phi$, student
  $q_\theta\!\leftarrow\!p_\phi$, weights $\lambda$, training steps $N_1$
\State \textbf{Output:} few-step student $q_\theta$, used to initialize Stage~2
\For{$N_1$ training steps}
  \State sample an example from $\mathcal{D}$ and two keyframes $x_t,x_{t'}$ with $t<t'$
  \State $\mathcal{L}_{\mathrm{traj}}\gets$ forward KL to the cached teacher prediction at the positions committed between them
  \State $\mathcal{L}_{\mathrm{sc}}\gets\KL\big(\mathrm{sg}[q_\theta(\cdot\mid x_{t'})]\,\|\,q_\theta(\cdot\mid x_{t})\big)$
  \State $\mathcal{L}_{\mathrm{path}}\gets$ cross-entropy to the teacher's final tokens on a rebuilt intermediate state
  \State $\mathcal{L}_{\mathrm{gt}}\gets\mathrm{CE}$ on freshly masked data
  \State update $\theta$ with $\mathcal{L}_1$ (Eq.~\ref{eq:stage1})
\EndFor
\end{algorithmic}
\end{algorithm}

\paragraph{Sparse teacher predictions.} Storing the full distribution at every cached position is
too costly, so we keep the teacher's top-$R$ log-probabilities and one residual probability for the
rest of the vocabulary. Let $p_1,\dots,p_R$ be the corresponding teacher probabilities and
$p_{R+1}=1-\sum_k p_k$ the residual probability. Let $q_1,\dots,q_R$ be the student's probabilities
of the same tokens, with $q_{R+1}$ defined in the same way. Both models are then compared on the same
$R{+}1$ categories,
\begin{equation}\small
\KL\big(p^{(R)}\,\big\|\,q^{(R)}\big)
=\sum_{k=1}^{R}p_k\log\frac{p_k}{q_k}\;+\;p_{R+1}\log\frac{p_{R+1}}{q_{R+1}}.
\label{eq:topk}
\end{equation}

\paragraph{Stage-1 terms.} Let $x_t,x_{t'}$ with $t<t'$ be two keyframes, so $x_t$ is the
earlier one and has more masked positions. Let $\mathcal{U}$ be the positions the teacher commits
between them, and $\mathcal{M}(x_{t'})$ the positions still masked at the later one. Then
\begin{equation}\small
\mathcal{L}_{\mathrm{traj}}=\E\Big[\tfrac1{|\mathcal{U}|}\!\sum_{i\in\mathcal{U}}\KL\big(p^{(R),i}_\phi\,\|\,q^{(R),i}_\theta\big)\Big],
\qquad
\mathcal{L}_{\mathrm{sc}}=\E\Big[\tfrac1{|\mathcal{M}(x_{t'})|}\!\!\!\sum_{i\in\mathcal{M}(x_{t'})}\!\!\!
\KL\big(\mathrm{sg}[q^{\,i}_\theta(\cdot\mid x_{t'})]\,\|\,q^{\,i}_\theta(\cdot\mid x_{t})\big)\Big],
\label{eq:stage1-terms}
\end{equation}
where the student predicts on $x_t$ in both terms, and $p^{(R),i}$ is the sparse teacher prediction
at position $i$. For the other two terms, let $\tilde{x}$ be an intermediate state rebuilt from the
cached commit steps, with uncommitted positions $\mathcal{M}(\tilde{x})$. Let $\hat{y}_i$ be the token
the teacher finally commits at position $i$. Let $x$ be a randomly masked copy of a training output
$y$, with masking rate $\pi$ and output length $n$. Then
\begin{equation}\small
\mathcal{L}_{\mathrm{path}}=\E\Big[\tfrac1{|\mathcal{M}(\tilde{x})|}\!\!\!\sum_{i\in\mathcal{M}(\tilde{x})}\!\!\!
-\log q^{\,i}_\theta(\hat{y}_i\mid\tilde{x})\Big],
\qquad
\mathcal{L}_{\mathrm{gt}}=\E\Big[\tfrac1{n}\sum_{i\in\mathcal{M}(x)}\tfrac1{\pi}\,
\big(\!-\log q^{\,i}_\theta(y_i\mid x)\big)\Big].
\label{eq:stage1-rest}
\end{equation}
We weight by $1/\pi$ and divide by $n$ instead of the number of masked positions. This makes
$\mathcal{L}_{\mathrm{gt}}$ an unbiased estimate of the full masked cross-entropy.

\paragraph{Image gradient and its surrogate loss.} Distribution matching for continuous diffusion
backpropagates through a reparameterized sample. The gradient of the divergence then becomes the
difference between two scores \citep{yin2024dmd, yin2024dmd2}. Sampling discrete tokens is not
differentiable, so this path does not exist. Following Di[M]O \citep{zhu2025dimo} for masked image
generators, we specify the update directly instead of differentiating a divergence. The update
$\mathrm{sg}[r_\psi-p_\phi]$ in line~\ref{ln:img-g} of Algorithm~\ref{alg:stage2} is the categorical
counterpart of the score difference. For a single categorical $q=\mathrm{softmax}(z)$ and a target
$p$, the forward KL has gradient $\nabla_{\!z}\KL(p\,\|\,q)=q-p$. Our update replaces $q$ with the
student's own distribution at the scored state, estimated by $r_\psi$. We apply this update to
$z_\theta$ without differentiating either network by minimizing
\begin{equation}\small
\mathcal{L}^{\,\mathrm{img}}_{\mathrm{match}}
=\frac{1}{2|\mathcal{M}|}\big\|z_\theta-\mathrm{sg}\big[z_\theta-g\big]\big\|^{2},
\qquad g=\mathrm{sg}\big[r_\psi(\cdot\mid s)-p_\phi(\cdot\mid s)\big],
\label{eq:dmd-surrogate}
\end{equation}
whose gradient is $g/|\mathcal{M}|$ by construction. This keeps the term on the same per-position
scale as the other losses. We use the forward-KL direction because it encourages the student to
cover the teacher's support. Both scores are computed on the re-masked output instead of the
rollout state. On the rollout state, $r_\psi$ would collapse onto $q_\theta$, and the term would
reduce to per-position knowledge distillation.

\paragraph{Entropy-matched guidance.} Entropy increases with temperature, so we solve
Eq.~\ref{eq:em} for $\nu$ at each scored position by bisection on $[0.25,8]$ with fourteen iterations.
If the solution falls outside this range, we clip it to the nearest endpoint.

\paragraph{Preference term.} $\mathcal{L}_{\mathrm{pref}}$ follows T3D~\citep{zhang2026t3d}, which
applies direct discriminative optimization~\citep{zheng2025ddo} to few-step text dLLMs. It trains the
student to prefer a completion $y^{+}$ over a sample $y^{-}$ (line~\ref{ln:pref} of
Algorithm~\ref{alg:stage2}). The preferred completion $y^{+}$ is the teacher's completion of the
state. The dispreferred $y^{-}$ is sampled from a reference $q_{\mathrm{ref}}$, which is copied from
the student every $\tau$ updates. Let
$\Delta_\theta(y)=\frac{1}{|\mathcal{M}|}\sum_{i\in\mathcal{M}}\log\big(q_\theta^{\,i}(y_i\mid s)/q_{\mathrm{ref}}^{\,i}(y_i\mid s)\big)$
be the mean log-ratio between the student and the reference. Then
\begin{equation}\small
\mathcal{L}_{\mathrm{pref}}=-\log\sigma\big(\beta\,\Delta_\theta(y^{+})\big)
+\alpha\,\mathrm{softplus}\big(\beta\,\Delta_\theta(y^{-})\big),
\label{eq:pref}
\end{equation}
where $\sigma$ is the logistic function and $\alpha,\beta$ are scalar weights. For images, the term
is computed on the rollout state $x_t$ and its masked positions instead of $s$ and $\mathcal{M}$.

\paragraph{Inference.}\label{sec:inference} The student is trained on CFG targets, so it generates
images in a few decoding steps without an unconditional forward pass. For understanding, it decodes
block by block and reuses the KV cache exactly. This relies on the block-causal mask. The teacher
uses fully bidirectional attention and cannot reuse its cache.

\section{Experimental Details}
\label{sec:appendix}
\label{sec:appendix-exp}

\subsection{Training Data and Teacher Trajectories}
\label{sec:appendix-data}
\paragraph{Training data.} The training mixture has $51$K examples. T2I data ($11$K)
come from BLIP3o-60k~\citep{chen2025blip3o} and the Lumina-DiMOO training data.
I2I data ($22$K) come from UltraEdit~\citep{zhao2024ultraedit},
Subjects200K~\citep{tan2025ominicontrol}, Graph-200K~\citep{li2025visualcloze},
UniWorld-V1~\citep{lin2025uniworld} and the Lumina-DiMOO training data. MMU data ($13$K) come from
MAmmoTH-VL~\citep{guo2025mammothvl}, MultiUI~\citep{liu2025harnessing} and the Lumina-DiMOO training data. Inpainting pairs ($5$K)
are built from the T2I images by masking a centered region and serve only as extra training data. Each mini-batch contains one task,
sampled at the rates in Table~\ref{tab:hparams}.

\paragraph{Teacher trajectories.} We run Algorithm~\ref{alg:rollout} once over the mixture, and
every Stage-1 run reuses the resulting cache. Image trajectories use $T{=}64$ steps with
classifier-free guidance. The guidance strength $w_{\mathrm{task}}$ is $4$ on the prompt for T2I.
For editing, it is $2.5$ on the instruction and $4$ on the source image. Understanding trajectories
use $T{=}512$ steps without guidance. The full-step teacher in Section~\ref{sec:evalsetup} is
evaluated at $128$ image steps.

\subsection{Optimization and Cost}
\label{sec:appendix-cost}
Table~\ref{tab:hparams} gives the full training configuration. The two stages share the data
mixture, the batch size and the ground-truth anchor. They differ in the loss terms and the optimization
settings.

\begin{table}[h]
\caption{\textbf{Training Configuration.} The collision penalty applies only to the text branch.
The auxiliary network $r_\psi$ and the student reference $q_{\mathrm{ref}}$ are used only in Stage~2
and are discarded after training.}
\label{tab:hparams}
\begin{center}
\small
\begin{tabular}{lcc}
\toprule
 & Stage 1 (off-policy) & Stage 2 (on-policy) \\
\midrule
Optimizer & \multicolumn{2}{c}{AdamW ($\beta_1{=}0.9$, $\beta_2{=}0.95$)} \\
Learning rate (cosine) & $2{\times}10^{-5}\!\to\!2{\times}10^{-6}$ & $1{\times}10^{-5}\!\to\!1{\times}10^{-6}$ \\
Weight decay & $0.1$ & $0$ \\
Warmup updates & $50$ & $25$ \\
Global batch size & \multicolumn{2}{c}{$256$} \\
Attention mask (text / image) & \multicolumn{2}{c}{block-causal / bidirectional} \\
Task sampling prob.\ (T2I / MMU / I2I / Inpaint) & \multicolumn{2}{c}{$0.50$ / $0.25$ / $0.125$ / $0.125$} \\
Image resolution (T2I / I2I, inpaint) & \multicolumn{2}{c}{$1024{\times}1024$ / $512{\times}512$} \\
Text block size / output length & \multicolumn{2}{c}{$32$ / $512$} \\
Decoding steps per text block & \multicolumn{2}{c}{$4$} \\
Trajectory decoding steps (image / text) & \multicolumn{2}{c}{$64$ / $512$} \\
\midrule
Loss weights, Stage 1 & \multicolumn{2}{c}{$\lambda_{\mathrm{traj}}\,1.0$,\; $\lambda_{\mathrm{sc}}\,0.5$,\; $\lambda_{\mathrm{path}}\,0.2$,\; $\lambda_{\mathrm{gt}}\,0.1$} \\
Loss weights, Stage 2 & \multicolumn{2}{c}{$\lambda_{\mathrm{match}}\,0.5$,\; $\lambda_{\mathrm{col}}\,8.0$,\; $\lambda_{\mathrm{pref}}\,0.2$,\; $\lambda_{\mathrm{gt}}\,0.1$} \\
Keyframes per example (image / text) & $8$ / $64$ & N/A \\
Retained probabilities $R$ (image / text) & $512$ / $2048$ & N/A \\
Re-mask fraction range & N/A & $[0.2,0.9]$ \\
Entropy fraction $\rho$ & N/A & $0.7$ \\
Reference refresh $\tau$ (updates) & N/A & $20$ \\
Preference $\alpha$ / $\beta$ & N/A & $0.5$ / $0.1$ \\
\bottomrule
\end{tabular}
\end{center}
\end{table}

Table~\ref{tab:cost-train} reports the one-time training cost, and Table~\ref{tab:cost-infer}
reports the cost per output at the deployed budget. The wall-clock speedup differs from the NFE
reduction. Understanding speeds up by \gain{$21.2\times$}, more than its $8\times$ NFE reduction,
because the block-causal mask allows exact KV cache reuse (Appendix~\ref{sec:inference}).
Each block reuses the keys and values of the prompt and all earlier finished blocks, and recomputes
only the current block and the one just finished. Without the cache, the same student is only \gain{$6.2\times$} faster. T2I speeds up by
\gain{$18.2\times$}, less than its $32\times$ NFE reduction. Peak memory is the same for both
models, between $16$ and $21$\,GB.

\begin{table}[h]
\caption{\textbf{Training Cost.} One-time cost on 32 A100 GPUs. Trajectory collection runs once
over the training mixture, and every Stage-1 run reuses it.}
\label{tab:cost-train}
\begin{center}
\small
\begin{tabular}{lccc}
\toprule
Training & Wall-clock & GPU-hours & Models in memory \\
\midrule
Teacher trajectory collection & $50.5$\,h & $1{,}616$ & $p_\phi$ \\
Stage 1 (off-policy replay) & $29.0$\,h & $928$ & $q_\theta$ \\
Stage 2 (on-policy refinement) & $31.3$\,h & $1{,}002$ & $q_\theta$, $p_\phi$, $r_\psi$, $q_{\mathrm{ref}}$ \\
\midrule
Total & $110.8$\,h & $3{,}546$ & \\
\bottomrule
\end{tabular}
\end{center}
\end{table}

\begin{table}[h]
\caption{\textbf{Inference Cost per Output.} Measured on one A100-80GB at batch size 1 in
bfloat16. We time only the token-space sampler and report the mean of 10 runs after 2 warmup runs.
Parentheses give the student's speedup over the teacher. TPS is tokens per second.}
\label{tab:cost-infer}
\begin{center}
\small
\fit{%
\begin{tabular}{lcccccc}
\toprule
 & \multicolumn{2}{c}{NFE $\downarrow$} & \multicolumn{2}{c}{Latency (s) $\downarrow$} & \multicolumn{2}{c}{TPS $\uparrow$} \\
\cmidrule(lr){2-3} \cmidrule(lr){4-5} \cmidrule(lr){6-7}
Inference (per output) & Teacher & Student & Teacher & Student & Teacher & Student \\
\midrule
T2I & $128{\times}2{=}256$ & $8$ (\gain{$32\times$}) & $106.4$ & $5.84$ (\gain{$18.2\times$}) & $38.5$ & $701.4$ \\
I2I & $128{\times}3{=}384$ & $8$ (\gain{$48\times$}) & $79.6$ & $2.28$ (\gain{$34.9\times$}) & $12.9$ & $449.7$ \\
MMU & $512{\times}1{=}512$ & $64$ (\gain{$8\times$}) & $80.7$ & $3.81$ (\gain{$21.2\times$}) & $6.3$ & $134.5$ \\
\quad w/o KV cache & $512{\times}1{=}512$ & $64$ (\gain{$8\times$}) & $80.7$ & $13.12$ (\gain{$6.2\times$}) & $6.3$ & $39.0$ \\
\bottomrule
\end{tabular}}
\end{center}
\end{table}

\subsection{Baselines}
\label{sec:appendix-baselines}
All baselines are distilled from the same Lumina-DiMOO teacher with the same initialization,
training mixture and number of optimizer updates. They are evaluated with the same sampler at the
same decoding budgets, and each keeps its own published objective. Every method is trained once and
evaluated at all budgets from that single checkpoint, with no budget-specific retraining or tuning.
The only change is modality coverage. Di[M]O was designed for masked image generators, and CDLM and
T3D for text dLLMs. We apply each of them to the other modality as well.

\subsection{Benchmarks}
\label{sec:appendix-benchlist}
T2I evaluation covers GenEval~\citep{ghosh2023geneval}, DPG-Bench~\citep{hu2024ella},
UniGenBench~\citep{wang2025prefgrpo, wang2025unigenbench}, OneIG~\citep{chang2025oneig}
with English prompts and TIIF-Bench~\citep{wei2025tiif}. I2I is evaluated on the I2I benchmark and
Graph-200K~\citep{li2025visualcloze}. For MMU, the discriminative benchmarks are
POPE~\citep{li2023pope}, MME-P~\citep{fu2023mme}, MMBench~\citep{liu2024mmbench},
SEED~\citep{li2023seedbench} and MMMU~\citep{yue2024mmmu}. The open-ended benchmarks are
MM-Vet~\citep{yu2024mmvet}, Vibe-Eval~\citep{padlewski2024vibeeval},
LLaVA-Bench~\citep{liu2023visual}, COCO captioning~\citep{chen2015microsoft} and
DeCapBench~\citep{ye2025painting}. Each benchmark uses its official metric, except COCO captioning,
for which we define our own protocol in Appendix~\ref{sec:appendix-judge}.

\subsection{Open-Ended Judging Protocols}
\label{sec:appendix-judge}
All five open-ended benchmarks are scored by the same GPT-5.5 judge. Four of them keep their official
grading prompt and aggregation. For COCO captioning, we define our own protocol.

\begin{itemize}[leftmargin=*]
\setlength\itemsep{0em}
\item \textbf{MM-Vet}~\citep{yu2024mmvet}. We use the official soft-grading prompt. The judge sees
the question, the ground-truth answer and the prediction, and gives a correctness score in
$\{0.0,0.1,\dots,1.0\}$. We report the mean $\times 100$.
\item \textbf{Vibe-Eval}~\citep{padlewski2024vibeeval}. We use the official text-only grading
prompt. The judge rates the output against the reference on a $1$ to $5$ scale, and we rescale the
score $s$ to $0$ to $100$ by $25(s-1)$.
\item \textbf{LLaVA-Bench}~\citep{liu2023visual}. We use the standard pairwise protocol. The judge
sees the image caption instead of the image and scores both the model answer and the reference GPT-4
answer from $1$ to $10$. We report the relative score, which is the model's mean score divided by
the reference's mean score, times $100$.
\item \textbf{COCO captioning}~\citep{chen2015microsoft}. This protocol is ours. The judge sees the
image and rates each caption on accuracy, completeness and coherence from $1$ to $5$. It is told to
penalize hallucinated objects and not to reward length. The five reference captions are given as
extra context, not as ground truth. Scores are rescaled to $0$ to $100$ by $25(s-1)$. We use a fixed
subset of $1000$ validation images.
\item \textbf{DeCapBench}~\citep{ye2025painting}. We use the official DCSCORE pipeline. It splits a
caption into atomic facts, matches them against the oracle facts and checks each one against the
image. The score is the mean of two F1 scores, one over all atomic facts and one over descriptive facts
only, $\times 100$.
\end{itemize}

The discriminative benchmarks use no judge. For yes/no benchmarks, we search the output for a
standalone ``yes'' or ``no''. For multiple-choice benchmarks, we match a standalone option letter
and fall back to the full option string. This parsing makes the scores insensitive to the rest of
the output.

\section{Additional Results}
\label{sec:appendix-results}
This section extends Section~\ref{sec:exp} with results at smaller decoding budgets, results in each
category of the T2I benchmarks and more qualitative samples.

\subsection{Results at Smaller Decoding Budgets}
\label{sec:appendix-budgets}
Tables~\ref{tab:main-t2i-more} to~\ref{tab:mmu-open-more} report T2I, I2I and the two MMU groups at
the smaller decoding budgets that Section~\ref{sec:exp} omits. GenEval at all budgets is in
Table~\ref{tab:geneval}. On T2I and MMU, the results follow the same pattern as at the larger budgets.
On T2I, Omni-Diffusion-Distill leads every benchmark at every budget, and its margin generally grows as the
budget shrinks. On discriminative MMU, the scores stay flat down to $16$ steps, and
Omni-Diffusion-Distill is the best distilled method on all five benchmarks. On open-ended MMU, the
scores fall for every method. Omni-Diffusion-Distill keeps the largest margin on COCO captioning,
while T3D is higher on LLaVA-Bench at $32$ and $16$ steps. On I2I, Omni-Diffusion-Distill leads Add at
both budgets and Graph-200K at $4$ steps. On the I2I benchmark Overall, T3D is higher at both budgets and Di[M]O at $4$ steps, by at
most $0.07$. T3D is also higher on Graph-200K at $2$ steps.

\begin{table}[h]
\caption{\textbf{Quantitative Results on T2I at Smaller Decoding Budgets.} The format follows Table~\ref{tab:main-t2i}.
GenEval at all budgets is in Table~\ref{tab:geneval}.}
\label{tab:main-t2i-more}
\begin{center}
\fit{%
\small
\begin{tabular}{lccccc}
\toprule
Method & NFE $\downarrow$ & DPG-Bench $\uparrow$ & UniGenBench $\uparrow$ & OneIG $\uparrow$ & TIIF-Bench $\uparrow$ \\
\midrule
Teacher (full-step) & $128{\times}2$ & 83.8 & 0.722 & 0.445 & 0.781 \\
\midrule
\multicolumn{6}{l}{\emph{4 image decoding steps}} \\
T3D & $4{\times}1$ & 70.6 & 0.630 & 0.283 & 0.641 \\
Di[M]O & $4{\times}1$ & \snd{78.0} & \snd{0.689} & \snd{0.296} & \snd{0.716} \\
CDLM & $4{\times}1$ & 51.3 & 0.450 & 0.212 & 0.547 \\
\textbf{Omni-Diffusion-Distill} & $4{\times}1$ & \bst{83.1} & \bst{0.724} & \bst{0.368} & \bst{0.759} \\
Teacher (few-step) & $4{\times}2$ & 63.3 & 0.559 & 0.253 & 0.598 \\
\midrule
\multicolumn{6}{l}{\emph{2 image decoding steps}} \\
T3D & $2{\times}1$ & \snd{55.9} & 0.524 & 0.214 & 0.591 \\
Di[M]O & $2{\times}1$ & 55.7 & \snd{0.564} & \snd{0.225} & \snd{0.603} \\
CDLM & $2{\times}1$ & 13.5 & 0.119 & 0.099 & 0.179 \\
\textbf{Omni-Diffusion-Distill} & $2{\times}1$ & \bst{73.8} & \bst{0.677} & \bst{0.277} & \bst{0.654} \\
Teacher (few-step) & $2{\times}2$ & 37.0 & 0.344 & 0.174 & 0.484 \\
\bottomrule
\end{tabular}}
\vspace{-1em}
\end{center}
\end{table}

\begin{table}[h]
\caption{\textbf{Quantitative Results on I2I at Smaller Decoding Budgets.} The format follows Table~\ref{tab:main-i2i}.}
\label{tab:main-i2i-more}
\begin{center}
\fit{%
\begin{tabular}{lc ccccc c}
\toprule
\multirow{2}{*}{Method} & \multirow{2}{*}{NFE $\downarrow$} & \multicolumn{5}{c}{\textbf{I2I Benchmark}} & \textbf{Graph-200K} \\
\cmidrule(lr){3-7} \cmidrule(lr){8-8}
& & Add $\uparrow$ & Compose $\uparrow$ & Adjust $\uparrow$ & Style $\uparrow$ & Overall $\uparrow$ & DINOv2 $\uparrow$ \\
\midrule
Teacher (full-step) & $128{\times}3$ & 3.62 & 1.94 & 2.29 & 3.89 & 2.66 & 0.799 \\
\midrule
\multicolumn{8}{l}{\emph{4 image decoding steps}} \\
T3D & $4{\times}1$ & \snd{2.20} & \snd{1.80} & \bst{2.49} & 3.74 & \bst{2.30} & \snd{0.773} \\
Di[M]O & $4{\times}1$ & 2.12 & 1.76 & 2.16 & \bst{3.78} & \snd{2.25} & 0.768 \\
CDLM & $4{\times}1$ & 1.33 & 1.07 & 1.39 & 2.92 & 1.53 & 0.513 \\
\textbf{Omni-Diffusion-Distill} & $4{\times}1$ & \bst{2.91} & \bst{1.84} & \snd{2.42} & \bst{3.78} & 2.24 & \bst{0.783} \\
Teacher (few-step) & $4{\times}3$ & 3.01 & 1.65 & 2.28 & 3.84 & 2.46 & 0.758 \\
\midrule
\multicolumn{8}{l}{\emph{2 image decoding steps}} \\
T3D & $2{\times}1$ & 2.23 & \bst{1.87} & \snd{2.32} & \bst{3.63} & \bst{2.26} & \bst{0.732} \\
Di[M]O & $2{\times}1$ & \snd{2.27} & 1.42 & 2.10 & 3.47 & 2.13 & 0.681 \\
CDLM & $2{\times}1$ & 1.00 & 1.00 & 1.00 & 1.41 & 1.15 & 0.111 \\
\textbf{Omni-Diffusion-Distill} & $2{\times}1$ & \bst{2.86} & \snd{1.58} & \bst{2.51} & \snd{3.50} & \snd{2.19} & \snd{0.710} \\
Teacher (few-step) & $2{\times}3$ & 2.40 & 1.51 & 2.11 & 3.46 & 2.19 & 0.668 \\
\bottomrule
\end{tabular}}
\vspace{-1em}
\end{center}
\end{table}

\begin{table}[h]
\caption{\textbf{Quantitative Results on Discriminative MMU at Smaller Decoding Budgets.} The format follows Table~\ref{tab:mmu-disc}.}
\label{tab:mmu-disc-more}
\begin{center}
\small
\begin{tabular}{lc ccccc}
\toprule
Method & NFE $\downarrow$ & POPE $\uparrow$ & MME-P $\uparrow$ & MMBench $\uparrow$ & SEED $\uparrow$ & MMMU $\uparrow$ \\
\midrule
Teacher (full-step) & $512$ & 86.8 & 1552 & 91.2 & 85.3 & 58.3 \\
\midrule
\multicolumn{7}{l}{\emph{32 decoding steps}} \\
T3D & $32$ & 82.8 & 1387 & 60.1 & 79.0 & 41.4 \\
Di[M]O & $32$ & \snd{86.3} & \snd{1478} & 57.6 & 76.4 & 36.0 \\
CDLM & $32$ & 79.9 & 1318 & \snd{78.5} & \snd{79.3} & \snd{50.3} \\
\textbf{Omni-Diffusion-Distill} & $32$ & \bst{87.6} & \bst{1535} & \bst{89.2} & \bst{84.5} & \bst{56.2} \\
Teacher (few-step) & $32$ & 86.5 & 1571 & 91.0 & 84.9 & 58.8 \\
\midrule
\multicolumn{7}{l}{\emph{16 decoding steps}} \\
T3D & $16$ & 83.5 & 1419 & 60.9 & 77.2 & 41.6 \\
Di[M]O & $16$ & \snd{86.0} & \snd{1515} & 61.9 & \snd{77.7} & 43.3 \\
CDLM & $16$ & 85.3 & 1480 & \snd{71.8} & 73.2 & \snd{49.8} \\
\textbf{Omni-Diffusion-Distill} & $16$ & \bst{87.3} & \bst{1544} & \bst{89.2} & \bst{84.8} & \bst{56.0} \\
Teacher (few-step) & $16$ & 86.5 & 1551 & 90.6 & 85.2 & 58.4 \\
\bottomrule
\end{tabular}
\end{center}
\end{table}

\begin{table}[h]
\caption{\textbf{Quantitative Results on Open-Ended MMU at Smaller Decoding Budgets.} The format follows Table~\ref{tab:mmu-open}.}
\vspace{-1.2em}
\label{tab:mmu-open-more}
\begin{center}
\small
\fit{
\begin{tabular}{lc ccccc}
\toprule
Method & NFE $\downarrow$ & MM-Vet $\uparrow$ & Vibe-Eval $\uparrow$ & LLaVA-Bench $\uparrow$ & COCO $\uparrow$ & DeCapBench $\uparrow$ \\
\midrule
Teacher (full-step) & $512$ & 21.3 & 12.36 & 25.1 & 67.2 & 44.5 \\
\midrule
\multicolumn{7}{l}{\emph{32 decoding steps}} \\
T3D & $32$ & 14.9 & \snd{3.90} & \bst{17.2} & 13.5 & 8.9 \\
Di[M]O & $32$ & 13.2 & 3.53 & 13.2 & 24.4 & 10.5 \\
CDLM & $32$ & \snd{15.6} & 3.90 & 13.3 & \snd{31.4} & \snd{10.6} \\
\textbf{Omni-Diffusion-Distill} & $32$ & \bst{16.8} & \bst{6.41} & \snd{13.7} & \bst{40.0} & \bst{13.2} \\
Teacher (few-step) & $32$ & 15.8 & 3.62 & 12.2 & 17.3 & 10.3 \\
\midrule
\multicolumn{7}{l}{\emph{16 decoding steps}} \\
T3D & $16$ & 11.6 & \snd{3.25} & \bst{13.8} & 8.9 & 7.5 \\
Di[M]O & $16$ & 11.3 & 2.70 & 12.2 & 15.1 & \snd{8.5} \\
CDLM & $16$ & \snd{14.1} & 3.07 & 12.2 & \snd{18.1} & 5.9 \\
\textbf{Omni-Diffusion-Distill} & $16$ & \bst{15.0} & \bst{4.09} & \snd{13.1} & \bst{26.0} & \bst{10.4} \\
Teacher (few-step) & $16$ & 13.4 & 2.70 & 12.0 & 12.4 & 8.9 \\
\bottomrule
\end{tabular}}
\vspace{-1em}
\end{center}
\end{table}

\subsection{Results in Each Category of the T2I Benchmarks}
\label{sec:appendix-bench}
Tables~\ref{tab:geneval}, \ref{tab:dpg}, \ref{tab:unigen}, \ref{tab:oneig} and~\ref{tab:tiif}
report GenEval, DPG-Bench, UniGenBench, OneIG and TIIF-Bench in each category, following the official
protocol of each benchmark.

At $8$ decoding steps, Omni-Diffusion-Distill leads every overall score. Its largest margins are in
categories where several parts of a prompt must be satisfied at once, namely Two Obj.\ and Position on
GenEval, Other on DPG-Bench, Relation and Compound on UniGenBench, and Real-world on TIIF-Bench. It also holds up
on long prompts, with $0.764$ on TIIF-Bench Long and $0.772$ on Short, while the strongest baseline drops
from $0.719$ to $0.691$. In several categories it matches or exceeds the full-step teacher, including
DPG-Bench Other ($86.8$ against $85.2$), UniGenBench Relation ($0.828$ against $0.792$) and Compound ($0.778$
against $0.740$). The remaining gap lies mainly in text rendering, where OneIG Text is $0.259$ against
the teacher's $0.542$, and in the holistic DPG-Bench Global category.

\begin{table}[h]
\caption{\textbf{Results in Each Category of GenEval.} Overall is the mean of the six categories. ImageReward~\citep{xu2023imagereward} and Aesthetic are image quality metrics computed on the same generated images.}
\label{tab:geneval}
\begin{center}
\setlength{\tabcolsep}{3pt}
\fit{%
\begin{tabular}{lcccccccccc}
\toprule
Method & NFE $\downarrow$ & \hd{Single\\Obj.} & \hd{Two\\Obj.} & Counting & Colors & Position & Attribute & Overall $\uparrow$ & \hd{Image\\Reward}~$\uparrow$ & Aesthetic $\uparrow$ \\
\midrule
Teacher (full-step) & $128{\times}2$ & 0.984 & 0.934 & 0.822 & 0.896 & 0.825 & 0.743 & 0.867 & 1.33 & 3.65 \\
\midrule
\multicolumn{11}{l}{\emph{8 image decoding steps}} \\
T3D & $8{\times}1$ & 0.947 & 0.846 & 0.700 & 0.814 & 0.770 & 0.573 & 0.775 & 0.97 & 3.55 \\
Di[M]O & $8{\times}1$ & \snd{0.972} & \snd{0.848} & \bst{0.747} & \snd{0.840} & \snd{0.772} & \bst{0.677} & \snd{0.810} & \snd{1.12} & \snd{3.61} \\
CDLM & $8{\times}1$ & 0.806 & 0.518 & 0.503 & 0.585 & 0.403 & 0.310 & 0.521 & 0.29 & 3.33 \\
\textbf{Omni-Diffusion-Distill} & $8{\times}1$ & \bst{0.984} & \bst{0.891} & \snd{0.713} & \bst{0.864} & \bst{0.835} & \bst{0.677} & \bst{0.828} & \bst{1.25} & \bst{3.71} \\
Teacher (few-step) & $8{\times}2$ & 0.978 & 0.811 & 0.672 & 0.832 & 0.743 & 0.670 & 0.784 & 0.98 & 3.61 \\
\midrule
\multicolumn{11}{l}{\emph{4 image decoding steps}} \\
T3D & $4{\times}1$ & \snd{0.966} & \snd{0.770} & \bst{0.709} & \snd{0.809} & \snd{0.700} & \snd{0.562} & \snd{0.753} & \snd{0.85} & 3.50 \\
Di[M]O & $4{\times}1$ & 0.944 & 0.747 & \snd{0.647} & 0.795 & 0.665 & 0.555 & 0.726 & 0.82 & \snd{3.52} \\
CDLM & $4{\times}1$ & 0.622 & 0.164 & 0.228 & 0.465 & 0.128 & 0.128 & 0.289 & -0.73 & 3.04 \\
\textbf{Omni-Diffusion-Distill} & $4{\times}1$ & \bst{0.969} & \bst{0.851} & 0.641 & \bst{0.878} & \bst{0.807} & \bst{0.640} & \bst{0.798} & \bst{1.15} & \bst{3.69} \\
Teacher (few-step) & $4{\times}2$ & 0.794 & 0.376 & 0.356 & 0.614 & 0.420 & 0.270 & 0.472 & -0.02 & 3.47 \\
\midrule
\multicolumn{11}{l}{\emph{2 image decoding steps}} \\
T3D & $2{\times}1$ & \snd{0.803} & \snd{0.333} & \snd{0.378} & \snd{0.596} & \snd{0.347} & \snd{0.207} & \snd{0.444} & \snd{-0.11} & \snd{3.12} \\
Di[M]O & $2{\times}1$ & 0.694 & 0.232 & 0.281 & 0.503 & 0.217 & 0.125 & 0.342 & -0.58 & 2.96 \\
CDLM & $2{\times}1$ & 0.100 & 0.000 & 0.003 & 0.053 & 0.000 & 0.000 & 0.026 & -2.11 & 2.86 \\
\textbf{Omni-Diffusion-Distill} & $2{\times}1$ & \bst{0.884} & \bst{0.505} & \bst{0.419} & \bst{0.710} & \bst{0.485} & \bst{0.318} & \bst{0.553} & \bst{0.25} & \bst{3.38} \\
Teacher (few-step) & $2{\times}2$ & 0.466 & 0.051 & 0.084 & 0.285 & 0.077 & 0.020 & 0.164 & -1.27 & 3.22 \\
\bottomrule
\end{tabular}}
\vspace{-1.5em}
\end{center}
\end{table}

\begin{table}[h]
\caption{\textbf{Results in Each Category of DPG-Bench.} Overall is the official DPG-Bench score computed over all prompts.}
\label{tab:dpg}
\begin{center}
\fit{%
\begin{tabular}{lccccccc}
\toprule
Method & NFE $\downarrow$ & Global & Entity & Attribute & Relation & Other & Overall $\uparrow$ \\
\midrule
Teacher (full-step) & $128{\times}2$ & 81.5 & 90.5 & 88.2 & 93.0 & 85.2 & 83.8 \\
\midrule
\multicolumn{8}{l}{\emph{8 image decoding steps}} \\
T3D & $8{\times}1$ & 69.0 & 79.2 & 79.7 & 85.6 & 73.6 & 69.1 \\
Di[M]O & $8{\times}1$ & \bst{84.2} & \snd{87.6} & \snd{86.5} & \snd{92.0} & \snd{76.8} & \snd{80.8} \\
CDLM & $8{\times}1$ & 71.4 & 77.0 & 77.6 & 86.2 & 64.8 & 66.2 \\
\textbf{Omni-Diffusion-Distill} & $8{\times}1$ & \snd{77.8} & \bst{89.8} & \bst{87.7} & \bst{92.9} & \bst{86.8} & \bst{83.0} \\
Teacher (few-step) & $8{\times}2$ & 77.8 & 85.9 & 84.7 & 91.8 & 75.6 & 78.6 \\
\midrule
\multicolumn{8}{l}{\emph{4 image decoding steps}} \\
T3D & $4{\times}1$ & 72.6 & 80.0 & 80.9 & 88.0 & \snd{75.2} & 70.6 \\
Di[M]O & $4{\times}1$ & \snd{79.6} & \snd{85.7} & \snd{85.2} & \snd{91.0} & 67.6 & \snd{78.0} \\
CDLM & $4{\times}1$ & 64.7 & 62.3 & 67.2 & 80.5 & 40.4 & 51.3 \\
\textbf{Omni-Diffusion-Distill} & $4{\times}1$ & \bst{81.8} & \bst{89.7} & \bst{88.1} & \bst{92.6} & \bst{84.4} & \bst{83.1} \\
Teacher (few-step) & $4{\times}2$ & 71.7 & 72.6 & 77.5 & 86.5 & 56.0 & 63.3 \\
\midrule
\multicolumn{8}{l}{\emph{2 image decoding steps}} \\
T3D & $2{\times}1$ & 68.1 & 67.7 & \snd{72.8} & 82.4 & \snd{52.8} & \snd{55.9} \\
Di[M]O & $2{\times}1$ & \snd{74.5} & \snd{67.8} & 72.2 & \snd{83.2} & 45.6 & 55.7 \\
CDLM & $2{\times}1$ & 52.3 & 20.0 & 40.0 & 44.3 & 7.6 & 13.5 \\
\textbf{Omni-Diffusion-Distill} & $2{\times}1$ & \bst{79.0} & \bst{82.1} & \bst{82.1} & \bst{89.7} & \bst{71.6} & \bst{73.8} \\
Teacher (few-step) & $2{\times}2$ & 64.7 & 49.7 & 64.0 & 76.8 & 33.2 & 37.0 \\
\bottomrule
\end{tabular}}
\end{center}
\end{table}
\begin{table}[h]
\caption{\textbf{Results in Each Category of UniGenBench.} Overall is the mean of the ten categories. World Know.\ is world knowledge, and Logic is logical reasoning.}
\label{tab:unigen}
\begin{center}
\setlength{\tabcolsep}{3pt}
\fit{%
\begin{tabular}{lcccccccccccc}
\toprule
Method & NFE & Style & \hd{World\\Know.} & Attribute & Action & Relation & Logic & Grammar & Compound & Layout & Text & Overall \\
\midrule
Teacher (full-step) & $128{\times}2$ & 0.936 & 0.870 & 0.876 & 0.735 & 0.792 & 0.432 & 0.701 & 0.740 & 0.873 & 0.267 & 0.722 \\
\midrule
\multicolumn{13}{l}{\emph{8 image decoding steps}} \\
T3D & $8{\times}1$ & 0.850 & 0.752 & 0.788 & 0.656 & 0.682 & 0.366 & 0.682 & 0.651 & 0.808 & \bst{0.163} & 0.640 \\
Di[M]O & $8{\times}1$ & \bst{0.943} & \snd{0.851} & \snd{0.865} & \snd{0.740} & \snd{0.768} & \snd{0.435} & \bst{0.713} & \snd{0.733} & \snd{0.859} & \snd{0.095} & \snd{0.700} \\
CDLM & $8{\times}1$ & 0.798 & 0.644 & 0.664 & 0.554 & 0.561 & 0.293 & 0.595 & 0.543 & 0.674 & 0.035 & 0.536 \\
\textbf{Omni-Diffusion-Distill} & $8{\times}1$ & \snd{0.901} & \bst{0.869} & \bst{0.873} & \bst{0.752} & \bst{0.828} & \bst{0.453} & \snd{0.694} & \bst{0.778} & \bst{0.875} & 0.082 & \bst{0.710} \\
Teacher (few-step) & $8{\times}2$ & 0.917 & 0.810 & 0.824 & 0.709 & 0.753 & 0.424 & 0.677 & 0.737 & 0.846 & 0.067 & 0.676 \\
\midrule
\multicolumn{13}{l}{\emph{4 image decoding steps}} \\
T3D & $4{\times}1$ & 0.887 & 0.737 & 0.783 & 0.654 & 0.674 & 0.361 & 0.676 & 0.641 & 0.794 & \bst{0.096} & 0.630 \\
Di[M]O & $4{\times}1$ & \snd{0.923} & \snd{0.834} & \snd{0.849} & \snd{0.727} & \snd{0.776} & \snd{0.438} & \bst{0.716} & \snd{0.745} & \snd{0.847} & 0.034 & \snd{0.689} \\
CDLM & $4{\times}1$ & 0.711 & 0.478 & 0.563 & 0.434 & 0.498 & 0.255 & 0.511 & 0.457 & 0.586 & 0.004 & 0.450 \\
\textbf{Omni-Diffusion-Distill} & $4{\times}1$ & \bst{0.940} & \bst{0.881} & \bst{0.889} & \bst{0.772} & \bst{0.833} & \bst{0.482} & \snd{0.706} & \bst{0.798} & \bst{0.870} & \snd{0.064} & \bst{0.724} \\
Teacher (few-step) & $4{\times}2$ & 0.851 & 0.628 & 0.690 & 0.544 & 0.632 & 0.319 & 0.579 & 0.622 & 0.709 & 0.013 & 0.559 \\
\midrule
\multicolumn{13}{l}{\emph{2 image decoding steps}} \\
T3D & $2{\times}1$ & \snd{0.818} & 0.585 & 0.657 & 0.502 & 0.578 & 0.285 & 0.584 & 0.545 & 0.674 & \snd{0.015} & 0.524 \\
Di[M]O & $2{\times}1$ & 0.790 & \snd{0.646} & \snd{0.669} & \snd{0.557} & \snd{0.636} & \snd{0.334} & \snd{0.658} & \snd{0.602} & \snd{0.736} & 0.008 & \snd{0.564} \\
CDLM & $2{\times}1$ & 0.351 & 0.054 & 0.171 & 0.061 & 0.115 & 0.042 & 0.184 & 0.095 & 0.115 & 0.000 & 0.119 \\
\textbf{Omni-Diffusion-Distill} & $2{\times}1$ & \bst{0.872} & \bst{0.793} & \bst{0.836} & \bst{0.725} & \bst{0.792} & \bst{0.446} & \bst{0.704} & \bst{0.787} & \bst{0.787} & \bst{0.024} & \bst{0.677} \\
Teacher (few-step) & $2{\times}2$ & 0.648 & 0.285 & 0.471 & 0.260 & 0.374 & 0.148 & 0.430 & 0.348 & 0.474 & 0.003 & 0.344 \\
\bottomrule
\end{tabular}}
\end{center}
\end{table}
\begin{table}[h]
\caption{\textbf{Results in Each Category of OneIG.} Overall is the mean of the five categories.}
\label{tab:oneig}
\begin{center}
\fit{%
\begin{tabular}{lccccccc}
\toprule
Method & NFE $\downarrow$ & Alignment & Text & Reasoning & Style & Diversity & Overall $\uparrow$ \\
\midrule
Teacher (full-step) & $128{\times}2$ & 0.817 & 0.542 & 0.281 & 0.400 & 0.185 & 0.445 \\
\midrule
\multicolumn{8}{l}{\emph{8 image decoding steps}} \\
T3D & $8{\times}1$ & 0.667 & \snd{0.122} & 0.155 & 0.365 & 0.218 & 0.306 \\
Di[M]O & $8{\times}1$ & \snd{0.780} & 0.095 & \snd{0.164} & \snd{0.375} & \snd{0.223} & \snd{0.327} \\
CDLM & $8{\times}1$ & 0.577 & 0.029 & 0.148 & 0.318 & \bst{0.330} & 0.280 \\
\textbf{Omni-Diffusion-Distill} & $8{\times}1$ & \bst{0.801} & \bst{0.259} & \bst{0.245} & \bst{0.393} & 0.206 & \bst{0.381} \\
Teacher (few-step) & $8{\times}2$ & 0.725 & 0.200 & 0.210 & 0.375 & 0.275 & 0.357 \\
\midrule
\multicolumn{8}{l}{\emph{4 image decoding steps}} \\
T3D & $4{\times}1$ & 0.652 & \snd{0.066} & 0.150 & 0.335 & \snd{0.210} & 0.283 \\
Di[M]O & $4{\times}1$ & \snd{0.744} & 0.021 & \snd{0.163} & \snd{0.345} & 0.209 & \snd{0.296} \\
CDLM & $4{\times}1$ & 0.420 & 0.003 & 0.122 & 0.273 & \bst{0.243} & 0.212 \\
\textbf{Omni-Diffusion-Distill} & $4{\times}1$ & \bst{0.803} & \bst{0.221} & \bst{0.229} & \bst{0.385} & 0.204 & \bst{0.368} \\
Teacher (few-step) & $4{\times}2$ & 0.531 & 0.010 & 0.147 & 0.317 & 0.259 & 0.253 \\
\midrule
\multicolumn{8}{l}{\emph{2 image decoding steps}} \\
T3D & $2{\times}1$ & 0.468 & \snd{0.007} & 0.131 & \bst{0.300} & \snd{0.166} & 0.214 \\
Di[M]O & $2{\times}1$ & \snd{0.519} & 0.002 & \snd{0.142} & \snd{0.294} & 0.166 & \snd{0.225} \\
CDLM & $2{\times}1$ & 0.123 & 0.000 & 0.099 & 0.169 & 0.104 & 0.099 \\
\textbf{Omni-Diffusion-Distill} & $2{\times}1$ & \bst{0.680} & \bst{0.028} & \bst{0.168} & 0.293 & \bst{0.213} & \bst{0.277} \\
Teacher (few-step) & $2{\times}2$ & 0.310 & 0.000 & 0.122 & 0.271 & 0.167 & 0.174 \\
\bottomrule
\end{tabular}}
\end{center}
\end{table}
\begin{table}[h]
\caption{\textbf{Results in Each Category of TIIF-Bench.} Short and Long are the overall scores on the two prompt lengths, and Overall is their mean. The three category columns are scored on short prompts.}
\label{tab:tiif}
\begin{center}
\fit{%
\begin{tabular}{lccccccc}
\toprule
Method & NFE $\downarrow$ & Short & Long & Basic & Advanced & Real-world & Overall $\uparrow$ \\
\midrule
Teacher (full-step) & $128{\times}2$ & 0.808 & 0.755 & 0.869 & 0.813 & 0.690 & 0.781 \\
\midrule
\multicolumn{8}{l}{\emph{8 image decoding steps}} \\
T3D & $8{\times}1$ & 0.670 & 0.655 & 0.828 & 0.719 & 0.493 & 0.663 \\
Di[M]O & $8{\times}1$ & \snd{0.719} & \snd{0.691} & \bst{0.852} & \snd{0.762} & \snd{0.552} & \snd{0.705} \\
CDLM & $8{\times}1$ & 0.616 & 0.582 & 0.795 & 0.688 & 0.470 & 0.599 \\
\textbf{Omni-Diffusion-Distill} & $8{\times}1$ & \bst{0.772} & \bst{0.764} & \snd{0.847} & \bst{0.806} & \bst{0.657} & \bst{0.768} \\
Teacher (few-step) & $8{\times}2$ & 0.723 & 0.676 & 0.853 & 0.771 & 0.545 & 0.699 \\
\midrule
\multicolumn{8}{l}{\emph{4 image decoding steps}} \\
T3D & $4{\times}1$ & 0.640 & 0.641 & 0.819 & 0.707 & 0.437 & 0.641 \\
Di[M]O & $4{\times}1$ & \snd{0.732} & \snd{0.700} & \bst{0.857} & \snd{0.794} & \snd{0.545} & \snd{0.716} \\
CDLM & $4{\times}1$ & 0.579 & 0.515 & 0.812 & 0.659 & 0.358 & 0.547 \\
\textbf{Omni-Diffusion-Distill} & $4{\times}1$ & \bst{0.784} & \bst{0.735} & \snd{0.856} & \bst{0.809} & \bst{0.653} & \bst{0.759} \\
Teacher (few-step) & $4{\times}2$ & 0.618 & 0.578 & 0.757 & 0.704 & 0.444 & 0.598 \\
\midrule
\multicolumn{8}{l}{\emph{2 image decoding steps}} \\
T3D & $2{\times}1$ & 0.573 & \snd{0.609} & \snd{0.790} & 0.649 & 0.321 & 0.591 \\
Di[M]O & $2{\times}1$ & \snd{0.623} & 0.583 & \bst{0.847} & \snd{0.717} & \snd{0.369} & \snd{0.603} \\
CDLM & $2{\times}1$ & 0.160 & 0.199 & 0.244 & 0.181 & 0.056 & 0.179 \\
\textbf{Omni-Diffusion-Distill} & $2{\times}1$ & \bst{0.654} & \bst{0.654} & 0.782 & \bst{0.731} & \bst{0.560} & \bst{0.654} \\
Teacher (few-step) & $2{\times}2$ & 0.509 & 0.459 & 0.643 & 0.629 & 0.317 & 0.484 \\
\bottomrule
\end{tabular}}
\end{center}
\end{table}

\subsection{Additional Qualitative Samples}
\label{sec:appendix-qual}
\begin{figure}[h]
\includegraphics[width=\linewidth]{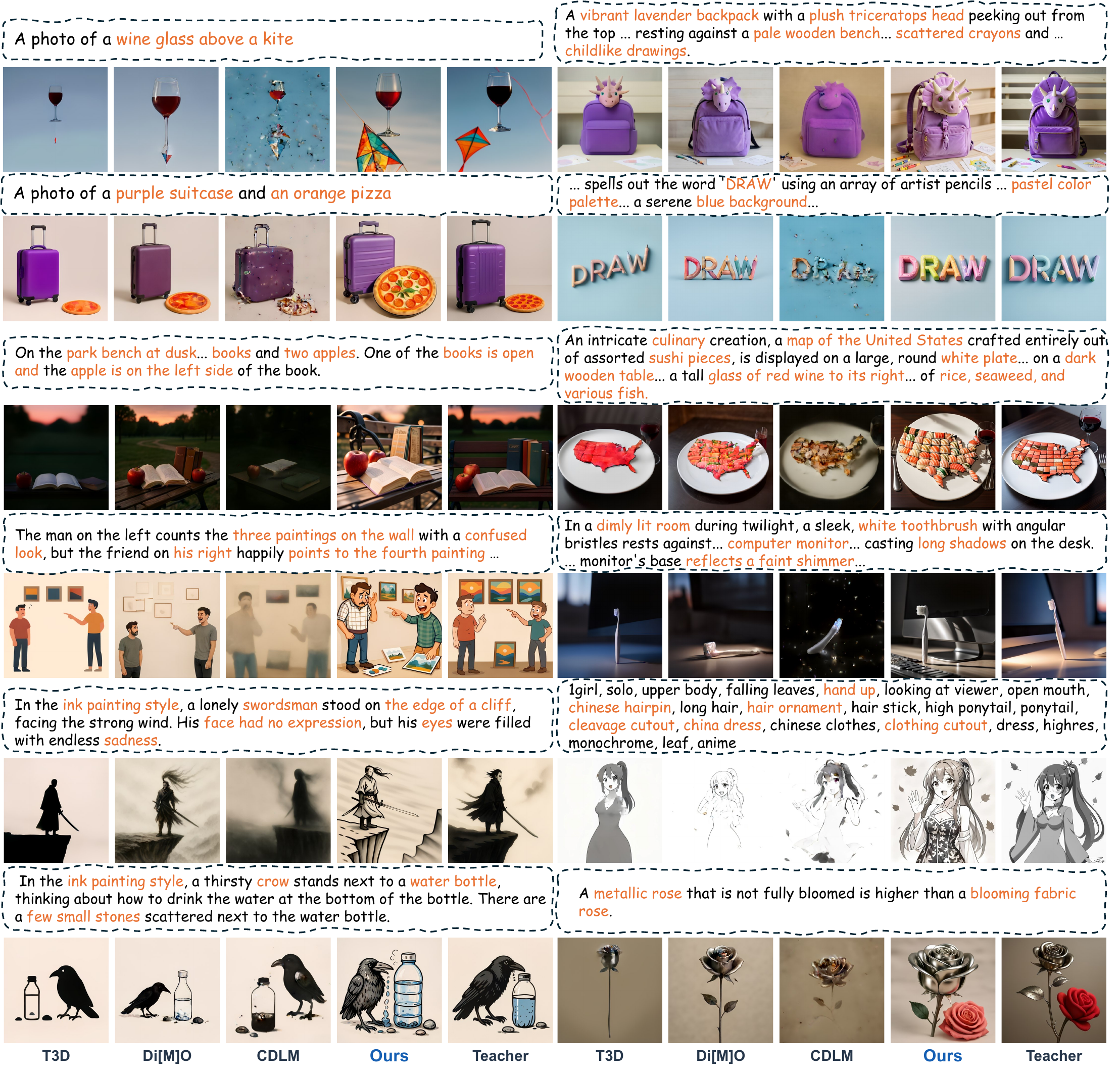}
\vspace{-2em}
\caption{\textbf{Additional qualitative comparison on T2I.}
All methods use the same prompts and seeds.
Distilled models use 8 decoding steps, and the teacher uses 128.}
\label{fig:qual-t2i-more}
\end{figure}

\begin{figure}[h]
\includegraphics[width=\linewidth]{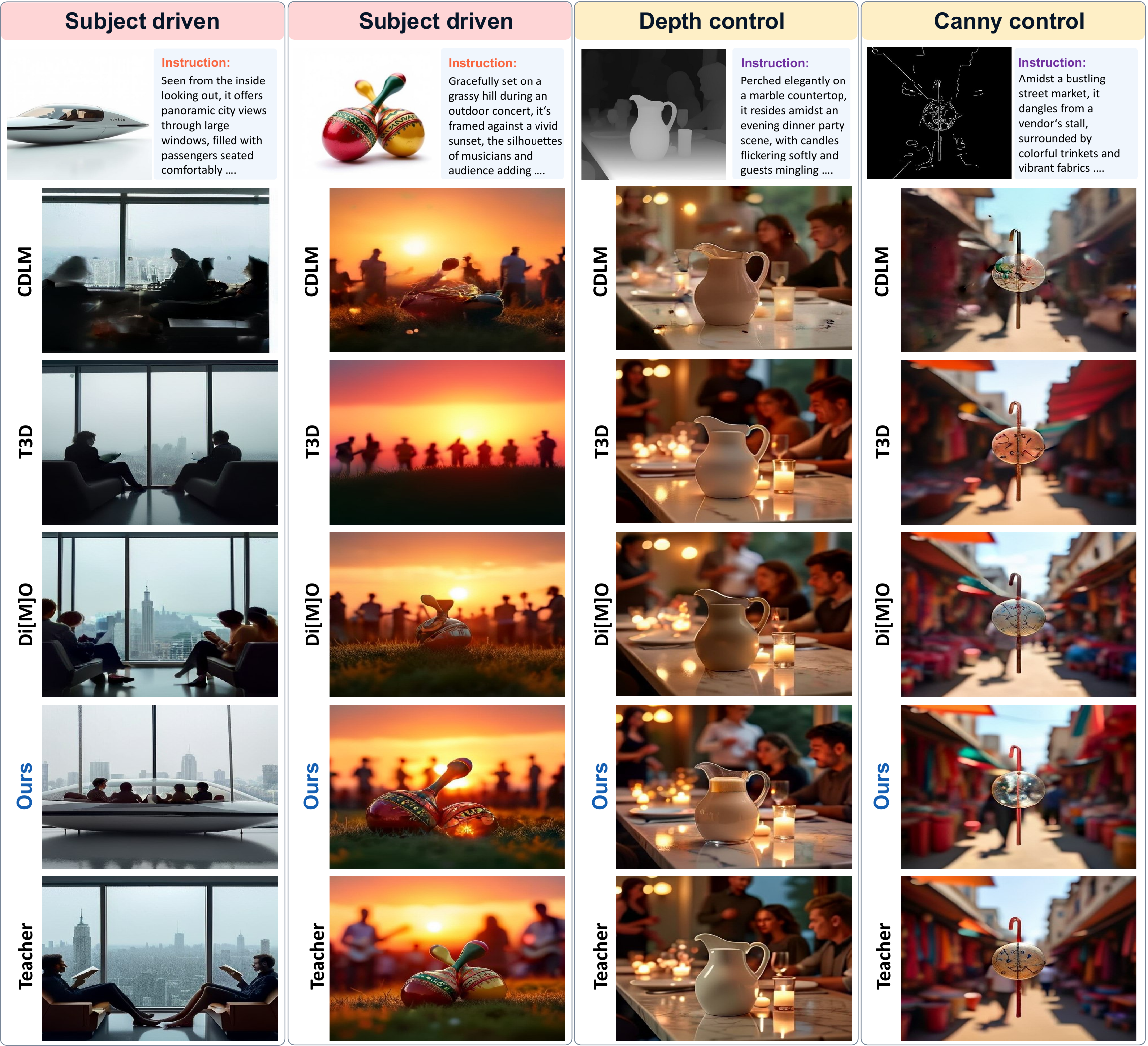}
\vspace{-2em}
\caption{\textbf{Additional qualitative comparison on controllable generation.}
All methods use the same conditioning inputs and prompts.
Distilled models use 8 decoding steps, and the teacher uses 128.}
\label{fig:qual-i2i-more}
\end{figure}

Figure~\ref{fig:qual-t2i-more} shows more T2I samples in the same format as Figure~\ref{fig:qual-t2i}, and
Figure~\ref{fig:qual-i2i-more} shows controllable generation samples.

\section{Analysis and Ablations}
\label{sec:appendix-eval}
This section checks whether the collision penalty and entropy-matched guidance work as intended, and which design choices matter
for the reported results.

\subsection{Why the Collision Penalty and Entropy-Matched Guidance Help}
\label{sec:appendix-analysis}

The collision penalty and entropy-matched guidance in Section~\ref{sec:stage2} address the two
challenges in Section~\ref{sec:intro}. These are repetition in parallel text decoding and entropy
collapse caused by fitting the sharpened CFG target in image distillation. Table~\ref{tab:analysis}
measures both directly.

\paragraph{Text degeneration metrics.}
We compute degeneration metrics on the outputs of the five open-ended understanding benchmarks and
average them across benchmarks. For a generated token sequence $y_1,\dots,y_n$, we report four
metrics.
\begin{itemize}
\item Rep-1 is the fraction of adjacent token pairs with $y_i=y_{i-1}$. It measures local token repetition.
\item Rep-4 is $1-|\{\text{unique 4-grams}\}|/|\{\text{4-grams}\}|$. It measures repeated phrases instead of repeated tokens.
\item Loop@5 is the fraction of outputs that contain at least five identical tokens in a row. It captures severe repetition loops.
\item Distinct-1 is the ratio of unique tokens to all tokens in an output. Higher values mean more diverse outputs.
\end{itemize}
Output lengths differ across models, so we compute Rep-4 and Distinct-1 on the first $64$ tokens of each output to compare them at equal length.

\begin{table}[t]
\caption{\textbf{Analysis of Text Repetition and Image Entropy.}
Left: degeneration metrics on the open-ended understanding benchmarks.
Right: predictive entropy of image tokens at the final decoding step.
Teacher rows use the raw CFG target, and student rows use the student's conditional prediction.}
\label{tab:analysis}
\begin{center}
\fit{%
\begin{tabular}{lcccc c}
\toprule
\multirow{2}{*}{Model} & \multicolumn{4}{c}{\textbf{Text degeneration}} & \textbf{Image entropy} \\
\cmidrule(lr){2-5} \cmidrule(lr){6-6}
& Rep-1 $\downarrow$ & Rep-4 $\downarrow$ & Loop@5 $\downarrow$ & Distinct-1 $\uparrow$ & $H_{\mathrm{final}}$ (nats) \\
\midrule
Teacher (full-step) & 0.020 & 0.043 & 0.006 & 0.704 & 2.13 \\
Teacher (few-step) & 0.473 & 0.459 & 0.744 & 0.313 & 2.28 \\
\midrule
\textbf{Omni-Diffusion-Distill} & \textbf{0.091} & \textbf{0.020} & \textbf{0.077} & \textbf{0.688} & 3.32 \\
\quad w/o collision penalty & 0.178 & 0.122 & 0.216 & 0.592 & \placeholder \\
\quad w/o entropy-matched guidance & \placeholder & \placeholder & \placeholder & \placeholder & 1.55 \\
\bottomrule
\end{tabular}}
\end{center}
\end{table}

\paragraph{Image predictive entropy.}
On the image side, classifier-free guidance produces a sharpened target, and we measure its effect
on the student directly. At the final decoding step, we compute the entropy of the predicted
distribution at every masked image position and average it over positions and samples to get
$H_{\mathrm{final}}$. For the teacher, we measure the raw CFG distribution
$\mathrm{softmax}(\ell^{\mathrm{cfg}})$ before Eq.~\ref{eq:em}. For the student, we measure its
conditional prediction, since the distilled student reproduces the guided distribution from a
conditional forward pass alone.

\paragraph{Results.}
On the text side, aggressive step reduction greatly increases degeneration. With the same teacher
weights, Loop@5 rises from $0.6\%$ with full-step decoding to $74.4\%$ at the student's decoding
budget. Omni-Diffusion-Distill reduces it to $7.7\%$. Without the collision penalty, it rises to
$21.6\%$, with higher Rep-1 and Rep-4 and lower Distinct-1. The collision penalty therefore directly
reduces the repetition caused by parallel text decoding.

On the image side, without entropy-matched guidance, the student's final entropy falls to $1.55$.
This is even below the $2.28$ of the raw CFG target of the few-step teacher at the same budget.
Entropy-matched guidance raises it to $3.32$, above the raw CFG target at both budgets.

\begin{figure}[h]
\centering
\includegraphics[width=0.62\linewidth]{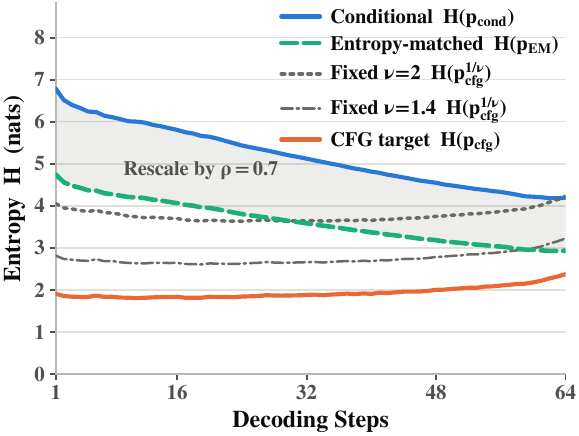}
\caption{\textbf{Entropy trajectories under adaptive and fixed-temperature scaling.} Entropy-matched
guidance targets $0.7$ times the conditional entropy along the 64-step teacher trajectory. The two fixed
temperatures produce lower entropy than this target early in decoding and higher entropy near the
end.}
\label{fig:entropy_temperature}
\end{figure}

Eq.~\ref{eq:em} solves for one temperature at each position and step. A simpler alternative is one
global temperature for the whole trajectory. Figure~\ref{fig:entropy_temperature} compares
entropy-matched guidance with fixed-temperature scaling. Entropy-matched guidance tracks the
decreasing conditional entropy at the fraction $\rho=0.7$ throughout decoding. A fixed temperature
cannot track this decrease. With either fixed temperature, the entropy falls below the entropy-matched target early in decoding
and exceeds it near the end. Table~\ref{tab:abl-em} shows that a fixed temperature also lowers the
benchmark scores.

Both measurements are consistent with the ablations in Tables~\ref{tab:abl-mmu} and~\ref{tab:abl-em}. The collision penalty brings its largest gain on
COCO captioning, and entropy-matched guidance brings its largest gain on the I2I benchmark. This matches what
each component acts on. The collision penalty acts on the probability that two scored positions in a
block commit the same token, and entropy-matched guidance acts on the entropy of the teacher
prediction.

\subsection{Ablations}
\label{sec:appendix-abl}
Table~\ref{tab:abl-stage} tests whether both training stages are needed, Table~\ref{tab:abl-mmu}
varies the understanding objective, and Table~\ref{tab:abl-em} varies the entropy fraction $\rho$
and compares entropy-matched guidance with a fixed temperature.

\begin{table}[h]
\caption{\textbf{Effect of the Two Training Stages.} Rows compare Stage~1 only, Stage~2 only
without the Stage~1 cold start, and the full two-stage training.}
\label{tab:abl-stage}
\begin{center}
\fit{%
\begin{tabular}{lccccc cc ccc}
\toprule
\multirow{2}{*}{Training} & \multicolumn{5}{c}{\textbf{Text-to-image}} & \multicolumn{2}{c}{\textbf{Image-to-image}} & \multicolumn{3}{c}{\textbf{Multimodal understanding}} \\
\cmidrule(lr){2-6} \cmidrule(lr){7-8} \cmidrule(lr){9-11}
 & GenEval $\uparrow$ & DPG-Bench $\uparrow$ & OneIG $\uparrow$ & TIIF-Bench $\uparrow$ & UniGenBench $\uparrow$ & I2I Benchmark $\uparrow$ & Graph-200K $\uparrow$ & MM-Vet $\uparrow$ & Vibe-Eval $\uparrow$ & COCO $\uparrow$ \\
\midrule
Stage 1 only & 0.740 & 68.2 & 0.282 & 0.585 & 0.585 & 2.10 & 0.765 & 17.0 & 7.34 & 36.1 \\
Stage 2 only & 0.771 & 81.8 & 0.333 & 0.719 & 0.684 & 2.15 & 0.769 & 19.4 & 7.62 & 56.5 \\
\midrule
\textbf{Ours} & \textbf{0.828} & \textbf{83.0} & \textbf{0.381} & \textbf{0.768} & \textbf{0.710} & \textbf{2.31} & \textbf{0.789} & \textbf{20.0} & \textbf{8.18} & \textbf{57.2} \\
\bottomrule
\end{tabular}}
\end{center}
\end{table}

\begin{table}[h]
\caption{\textbf{Ablation on the MMU Objective.} Rows vary the scored state, the form of the teacher prediction and the repetition term.}
\label{tab:abl-mmu}
\begin{center}
\fit{%
\begin{tabular}{llccccc}
\toprule
Scored state & Teacher prediction & Repetition term & MM-Vet $\uparrow$ & Vibe-Eval $\uparrow$ & COCO $\uparrow$ & DeCapBench $\uparrow$ \\
\midrule
\multirow{4}{*}{Sampler's own masked positions} & one-step marginal & none &  18.8 & 6.78 & 47.3 & 14.9 \\
\cmidrule(lr){2-7}
 & \multirow{3}{*}{sequential conditional} & none & 18.2 & 8.09 & 49.8 & 13.5 \\
 & & adjacent repeat & 18.9 & 7.62 & 49.3 & 14.4 \\
 & & pairwise collision & \textbf{20.0}& \textbf{8.18} & \textbf{57.2} & 19.6 \\
\midrule
\multirow{3}{*}{Re-masked committed tokens} & \multirow{3}{*}{sequential conditional} & none & 16.9 & 7.90 & 48.5 & 13.7 \\
 & & adjacent repeat & 16.8 & 7.53 & 49.5 & 14.6 \\
 & & pairwise collision & 19.4 &  8.18 & 53.4  & \textbf{21.2} \\
\bottomrule
\end{tabular}}
\end{center}
\end{table}

\paragraph{Both stages.} Both stages are needed. Stage~1 alone scores below the few-step teacher on every image
benchmark. Its role is to give Stage~2 a good starting point, not to serve as the final model.
Using Stage~1 as a cold start for Stage~2 improves every benchmark. The largest gain is $14.4\%$ on OneIG,
followed by $7.4\%$ on GenEval, $7.4\%$ on the I2I benchmark and $7.3\%$ on Vibe-Eval. The gains are smaller on MM-Vet, from $19.4$ to $20.0$, and on
COCO captioning, from $56.5$ to $57.2$. Without
Stage~2, the student scores $14.8$ points lower on DPG-Bench and $21.1$ lower on COCO captioning.

\paragraph{Understanding objective.} The scored state decides where the student is trained. With
the sampler's own masked positions, which is the state used in Section~\ref{sec:stage2-text}, we
score the positions that the current block has not yet committed at a random step of the student's
rollout. With re-masked committed tokens, we instead take positions the student has already
committed and mask them again. With the pairwise collision penalty, the sampler's own masked
positions give higher scores on MM-Vet and COCO captioning, the same score on Vibe-Eval and a lower
score on DeCapBench.

The teacher prediction and the repetition term have larger effects. The one-step marginal queries
the teacher once on the state, so the prediction at each position ignores the tokens committed at
the other positions. The sequential conditional lets the teacher commit these positions one at a
time, so each prediction sees the tokens the teacher has already committed. This change improves
Vibe-Eval from $6.78$ to $8.09$ and COCO captioning from $47.3$ to $49.8$, while MM-Vet and
DeCapBench drop slightly. It does not remove repetition by itself, as Section~\ref{sec:stage2-text}
predicts, because a per-position prediction has no explicit term for two positions in a block
committing the same token. The largest improvement comes from the pairwise collision penalty, which
acts on every pair of scored positions in a block. It raises COCO captioning to $57.2$ and DeCapBench
to $19.6$. An adjacent-repeat penalty acts only on neighboring positions and thus directly optimizes
Rep-1, but it brings almost no improvement.

\begin{table}[h]
\caption{\textbf{Ablation on Entropy-Matched Guidance.}
The entropy fraction $\rho$ is relative to the entropy of the conditional distribution at each scored
position. Off disables entropy-matched guidance and uses the raw CFG target. Fixed $\nu$ divides the
guided logits by one temperature at all positions and steps.}
\label{tab:abl-em}
\begin{center}
\fit{%
\begin{tabular}{lcccccccc}
\toprule
Setting
& GenEval $\uparrow$
& DPG-Bench $\uparrow$
& UniGenBench $\uparrow$
& OneIG $\uparrow$
& TIIF-Bench $\uparrow$
& ImageReward $\uparrow$
& Aesthetic $\uparrow$
& I2I Benchmark $\uparrow$ \\
\midrule
Off
& 0.798
& 81.5
& 0.690
& 0.343
& 0.723
& 1.23
& \textbf{3.73}
& 2.04 \\

$\rho=0.5$
& 0.815
& 79.0
& 0.704
& 0.377
& 0.745
& 1.19
& 3.65
& 2.20 \\

$\rho=0.7$ (Ours)
& \textbf{0.828}
& \textbf{83.0}
& \textbf{0.710}
& \textbf{0.381}
& \textbf{0.768}
& \textbf{1.25}
& 3.71
& \textbf{2.31} \\

$\rho=0.9$
& 0.176
& 67.5
& 0.562
& 0.260
& 0.549
& $-0.89$
& 3.33
& 2.00 \\
\midrule

Fixed $\nu=1.4$
& 0.475
& 78.6
& 0.670
& 0.304
& 0.707
& 0.72
& 3.52
& 2.03 \\

Fixed $\nu=2.0$
& 0.015
& 26.2
& 0.185
& 0.137
& 0.298
& $-2.11$
& 3.15
& 1.98 \\
\bottomrule
\end{tabular}}
\end{center}
\end{table}

\paragraph{Entropy fraction.} Among Off and the three values of $\rho$, $\rho=0.7$ is the best on
seven of the eight columns. The gain is largest on the I2I benchmark, where an over-confident student under-edits, with the
score rising from $2.04$ to $2.31$. The raw CFG target is higher only on Aesthetic, by $0.02$.
Softening further hurts. At $\rho=0.9$, generation quality drops sharply, with GenEval at $0.176$ and
a negative ImageReward. This suggests that too much softening weakens the guided target.

\paragraph{Fixed temperature.} Entropy-matched guidance solves for a temperature at each position
and step. A simpler alternative divides the guided logits by one fixed temperature, $\nu=1.4$ or
$\nu=2.0$. Both settings score below Off on every benchmark we evaluate, and the drop grows with
$\nu$. At $\nu=1.4$, GenEval falls from $0.798$ to $0.475$. At $\nu=2.0$, GenEval falls to $0.015$,
DPG-Bench to $26.2$ and ImageReward to $-2.11$. On the I2I benchmark, the two settings score $2.03$ and $1.98$, while
entropy-matched guidance scores $2.31$. A fixed temperature raises the entropy
above the target late in decoding (Figure~\ref{fig:entropy_temperature}), which likely weakens the
guided target. The gain of entropy-matched guidance thus comes from matching the entropy at each
position, not from softening the target alone.

\end{document}